\documentclass[pmlr,twocolumn,10pt]{jmlr}

\mlhtrack{proceedings}

\newif\iffinal
\finaltrue

\iffinal
  \ifmlhneedspmlr
    \jmlrvolume{346}
    \jmlryear{2026}
  \fi
  \ifmlhfindings \jmlrproceedings{}{ML4H 2026 - Findings Track}\fi
  \ifmlhdemo     \jmlrproceedings{}{ML4H 2026 - Demo Track}\fi
  \jmlrworkshop{Machine Learning for Health (ML4H) 2026}
\else
  \jmlrproceedings{}{Submitted to ML4H 2026: \mlhtrackname}
  \jmlrworkshop{Machine Learning for Health (ML4H) 2026}
\fi

\newcommand{\comblabel}{combined}

\newcommand{\coderepo}{\url{https://github.com/bronsonj98/recognition-delay}}

\usepackage{array}
\usepackage{booktabs}
\usepackage{graphicx}
\usepackage{multirow}
\usepackage{longtable}
\usepackage{needspace}
\hypersetup{pdftitle={The Label Is Not the Disease: Recognition Delay in Diagnosis Codes and Its Cost to EHR Case Finding}, pdfauthor={Moonseong Jeong, Prateek Anand, Simon A. Lee, Sungjoo Han, Zhengtong Liu, Sriram Sankararaman}}
\title[The Label Is Not the Disease]{The Label Is Not the Disease: Recognition Delay in Diagnosis Codes and Its Cost to EHR Case Finding}
\author{%
 \Name{Moonseong Jeong\nametag{$^{1}$}} \Email{bronsonj@cs.ucla.edu}\\
 \Name{Prateek Anand\nametag{$^{1}$}}\\
 \Name{Simon A. Lee\nametag{$^{2}$}}\\
 \Name{Sungjoo Han\nametag{$^{3}$}}\\
 \Name{Zhengtong Liu\nametag{$^{1}$}}\\
 \Name{Sriram Sankararaman\nametag{$^{1,2,4}$}} \Email{sriram@cs.ucla.edu}\\
 \addr $^{1}$Department of Computer Science, UCLA, Los Angeles, CA, USA\\
 \addr $^{2}$Department of Computational Medicine, David Geffen School of Medicine, UCLA, Los Angeles, CA, USA\\
 \addr $^{3}$Medical Informatics Home Area, Department of Bioinformatics, UCLA, Los Angeles, CA, USA\\
 \addr $^{4}$Department of Human Genetics, David Geffen School of Medicine, UCLA, Los Angeles, CA, USA
}

\begin{document}

\maketitle

\begin{abstract}
A condition can be present for years before its diagnosis enters the electronic health record, so machine learning models trained and evaluated on diagnosis codes may learn who is recognized rather than who has the disease. We measured this recognition delay in 186,449 UK Biobank participants whose blood was assayed at enrollment regardless of clinical need. Between 19\% (diabetes) and 62\% (anemia) of participants with values in the disease range had no diagnosis or treatment record, and for chronic kidney disease and anemia, recording caught up slowly. Recording was also selective, as it could be predicted from the pre-enrollment record. Anemia was recognized more often in the most deprived and in non-white participants after adjustment for severity. In case finding, the task of choosing which people without a diagnosis to test, gradient-boosting models trained on recorded diagnoses found about a quarter fewer people with disease-range values than models trained on the enrollment assays. Among anemia boosting models trained on different labels, the one that best predicted later diagnoses had the lowest AUROC for anemia-range hemoglobin at enrollment. Recorded diagnoses are therefore an incomplete and selective label for case finding, and later diagnoses are an imperfect target for evaluating it.
\end{abstract}

\paragraph*{Data and Code Availability} UK Biobank data are available to approved researchers through UK Biobank. Analysis code, including the drug lists and the rules that build the diagnosis-code lists, is available at \coderepo.

\paragraph*{Institutional Review Board (IRB)} This study used de-identified UK Biobank data under application 33127 and did not require IRB approval.

\section{Introduction}

A person can have glycated hemoglobin (HbA1c) in the diabetes range, reduced kidney function, or low hemoglobin for years before a diagnosis enters their electronic health record (EHR)\@. This lag creates two problems for clinical machine learning. A person without a diagnosis code may already have the condition, and a new code may mark the late recognition of a condition that was already present. Both problems matter most for case finding, the task of choosing which people without a diagnosis to test. A model trained on diagnosis codes may learn who is likely to be diagnosed, and an evaluation against later diagnoses may reward the same prediction.

Measuring what the record misses is difficult, because clinical laboratory tests are ordered selectively, depending on symptoms, prior diagnoses, and contact with care \citep{Agniel2018,Chang2022}. In UK Biobank, by contrast, blood from every participant was assayed at enrollment in 2006--2010 under the study protocol, without a clinician deciding who needed a test \citep{Sudlow2015}. We can therefore identify participants whose values were in the disease range regardless of their records, and follow their linked records to see whether and when a diagnosis was recorded.

We distinguish three quantities. A disease-range value is an enrollment assay result beyond a diagnostic threshold. It indicates biochemical abnormality at one time point and is not by itself a diagnosis, which also requires confirmation and clinical judgment. Clinical recognition is a clinician's awareness of the condition, which the records capture only in part, and a recorded diagnosis is a diagnosis code or relevant prescription in the linked records. In this paper, recognition means recorded recognition, the first such record after enrollment. A missing record does not by itself show that a condition was missed, since a clinician may judge that a borderline value does not need a diagnosis.

Studies in clinical records have compared laboratory evidence with coded diagnoses, using tests that clinicians ordered \citep{Schulz2021,Sisk2024}. The closest study is that of \citet{Young2022}, who measured the delay to diabetes diagnosis after a diabetes-range HbA1c at UK Biobank enrollment and evaluated existing screening scores against the same measurement. They counted an abnormal primary-care HbA1c or glucose value as a diagnosis, whereas we count only diagnosis codes and prescriptions, because these are what EHR labels are built from. This lets us show how often recording fails to follow an abnormal primary-care value. We also study kidney function and anemia, and we ask how recognition delay affects models trained and evaluated on recorded diagnoses. Other UK Biobank studies scored diabetes risk models against enrollment HbA1c \citep{He2021} or identified participants with low estimated glomerular filtration rate (eGFR) but no code for chronic kidney disease (CKD) \citep{Rodwell2026}.

In machine learning, the same problem appears as label bias. Training on a proxy label can distort whom a model selects \citep{Obermeyer2019,ZangerTishler2024}, and selective testing leaves untested patients labeled as negative \citep{Chang2022,ChangWiens2024,Balachandar2024}. Positive--unlabeled (PU) learning corrects for missing positive labels \citep{ElkanNoto2008,BekkerDavis2018}. These methods must assume how labels are recorded, and they are usually evaluated with simulated or partly simulated labels. \citet{BenitezAurioles2026} modeled unequal CKD testing and diagnosis in UK Biobank with a causal hidden Markov model that does not use kidney-function measurements, and corrected the resulting bias in a risk model. MILTON trains biomarker-based classifiers on recorded diagnoses to flag likely undiagnosed controls, and EDGAR trains on biomarker-defined phenotypes to estimate lifetime liability, both for genetic discovery \citep{Garg2024,DiCai2026}. We instead measure when disease-range values are recorded and what the choice of training label costs a case-finding model evaluated in unrecorded people. To our knowledge, no study has compared models trained on recorded diagnoses with models trained on enrollment assays in the same participants, or shown how evaluation against later diagnoses changes which model is chosen.

\begin{figure*}[t]
\floatconts{fig:design}{\caption{\textbf{Study design.} A, timeline for one participant; records more than a year before enrollment are not drawn to scale. The prior record ($R$) uses all records up to enrollment, and the features summarize the year before enrollment. Models predict at enrollment, when the assay ($D$) is measured, and $D$ is never an input. B, value of each training label by prior record and enrollment assay (--, not used). The incident label is 1 for participants recognized within the chosen horizon (1, 2, or 5 years). Models are evaluated in held-out participants with $R=0$, in whom $D=1$ marks a case.}}{\centering\includegraphics[width=\textwidth]{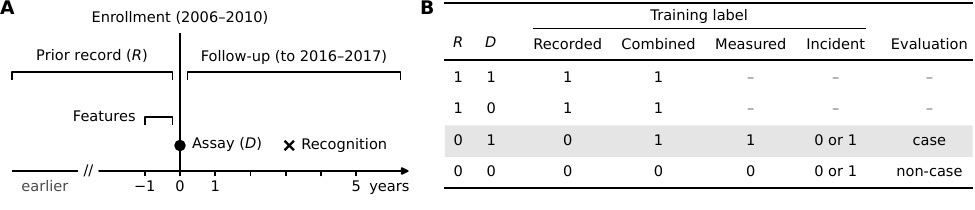}}
\end{figure*}

\begin{table*}[t]
\floatconts{tab:terms}{\caption{Definitions. Each applies to one condition at a time.}}{\centering\footnotesize
\begin{tabular}{@{}>{\raggedright\arraybackslash}p{0.17\textwidth}p{0.8\textwidth}@{}}
\toprule
Term & Definition \\
\midrule
Disease-range value ($D=1$) & Enrollment assay beyond the threshold. Diabetes range is HbA1c $\geq$48 mmol/mol. CKD range is creatinine and cystatin C eGFR both $<$60 mL/min/1.73 m$^2$. Anemia range is hemoglobin $<$12 g/dL in women or $<$13 g/dL in men. Participants with $D=1$ are assay-positive. \\\addlinespace[4pt]
Prior record ($R=1$) & A self-reported diagnosis or medication at enrollment, or a diagnosis code or relevant prescription in any earlier record. Unrecorded means $R=0$. \\\addlinespace[4pt]
Recognition & The first diagnosis code (primary-care, hospital, or death record) or relevant primary-care prescription after enrollment. \\\addlinespace[4pt]
Prior-year testing & A primary-care measurement of the same analyte in the year before enrollment, with creatinine used for kidney function. \\\addlinespace[4pt]
Abnormal primary-care value & A primary-care measurement of the same analyte after enrollment (a retest) that is beyond the threshold, with creatinine eGFR $<$60 used for kidney function. \\\addlinespace[4pt]
Training labels & The recorded label is $R$, and the \comblabel{} label is $R$ or $D$, both among all training participants. The measured label is $D$, and the incident label is recognition within 1, 2, or 5 years, both among training participants with $R=0$. \\\addlinespace[4pt]
Case-finding yield & Assay-positive participants (cases) found per 1,000 tested, when the highest-scoring 10\% of held-out unrecorded participants are tested. \\\addlinespace[4pt]
Recording propensity & The probability that an assay-positive participant has a prior record, given the 115 pre-enrollment features without code history ($X$). \\
\bottomrule
\end{tabular}}
\end{table*}

\paragraph{Contributions.}
\begin{itemize}\setlength\itemsep{0.1em}
\item \textbf{Incomplete and slow recording.} At enrollment, 19\% of participants with diabetes-range HbA1c, 32\% with CKD-range eGFR, and 62\% with anemia-range hemoglobin had no record of the condition. Recognition was slow for kidney function and anemia, often despite a later abnormal primary-care value, and new diabetes diagnoses often marked disease-range values present at enrollment (Sections~\ref{sec:recognition} and~\ref{sec:incident}).
\item \textbf{Selective recording.} Recording could be predicted from pre-enrollment features, and for anemia, recognition differed by deprivation and ethnicity in ways that severity did not explain (Section~\ref{sec:selective}).
\item \textbf{Consequences for models.} Gradient-boosting models trained on recorded diagnoses found 23--29\% fewer unrecorded participants with disease-range values than models trained on the enrollment assays, and they selected different participants for testing. Scoring models against later diagnoses instead of the enrollment assays favored a different anemia model (Sections~\ref{sec:models} to~\ref{sec:targets}).
\end{itemize}

\section{Study Design}\label{sec:design}

Figure~\ref{fig:design} shows when each quantity is measured and how each training label is defined, and Table~\ref{tab:terms} defines the terms used throughout.

\subsection{Cohort and Enrollment Assays}

UK Biobank enrolled 502,367 participants in 2006--2010 \citep{Sudlow2015}. Our cohort comprises the 186,449 participants with a continuous primary-care registration covering at least the year before enrollment, with records from a single general practice (GP) data supplier, so that diagnoses before and after enrollment can both be observed. Each analysis uses the participants with the required assays.

For each of the three conditions, $D=1$ when the enrollment assay is in the disease range (Table~\ref{tab:terms}). The HbA1c and hemoglobin thresholds are those of the World Health Organization \citep{WHOAnaemia2011,WHOHbA1c2011}. For kidney function we require both creatinine- and cystatin C-based eGFR below 60 mL/min/1.73 m$^2$, from the race-free 2021 CKD-EPI creatinine and 2012 cystatin C equations \citep{Inker2021,Inker2012}. Agreement between the two markers makes a low value more likely to reflect reduced kidney function, although a CKD diagnosis also requires chronicity \citep{KDIGO2024}.

\subsection{Prior Record and Recognition}

For each condition, $R=1$ when a diagnosis was recorded before enrollment, from self-reported diagnoses and medications at enrollment, UK Biobank first-occurrence dates (the earliest date of each condition in primary-care, hospital, death, or self-report data), primary-care diagnosis codes, and primary-care prescriptions of glucose-lowering drugs (diabetes) or iron (anemia). Kidney function uses diagnoses only. Prescriptions are matched by drug name, which excludes diabetes monitoring supplies, folic acid, and vitamin B12. Recognition is the earliest diagnosis code from primary-care, hospital, or death records, or the earliest relevant primary-care prescription, after enrollment. First-occurrence dates that come only from self-report are not used for recognition, because participants report them retrospectively. Appendix~\ref{app:defs} describes the code and drug lists. Follow-up ends at the earlier of the end of the primary-care extract (2016--2017) and the end of hospital records, with death as a competing event. Median follow-up was 7.3 years.

\subsection{Analyses of Recognition}\label{sec:recog_methods}

We estimate cumulative recognition among unrecorded assay-positive participants ($D=1$, $R=0$) with the Aalen--Johansen estimator \citep{Aalen1978} and 1,000 bootstrap resamples. For participants never recognized, we classify follow-up into three categories, at least one abnormal primary-care value, normal retests only, and no retest (for kidney function, an abnormal value is creatinine eGFR below 60, because cystatin C is rarely measured in primary care). To estimate recognition within a year of an abnormal value, we start a new clock at the first such value, and separately at a second abnormal value at least 90 days later. Analyses of primary-care values use practices in England and Wales, because many Scottish results are stored in a format that cannot be fully decoded.

We compare five-year recognition between groups defined by sex, Townsend deprivation quintile, ethnicity, age, and prior-year testing, with paired bootstrap differences and Holm adjustment within each condition. To ask whether differences arise because some groups have milder values, we fit Fine--Gray models \citep{FineGray1999} with and without a spline of severity, the distance of the enrollment value beyond the threshold, adjusting for age, sex, deprivation, ethnicity, GP data supplier, enrollment year, and prior-year testing. These comparisons are descriptive. Finally, we classify participants first recognized after enrollment by their enrollment value, and estimate how often disease-range values persist among participants remeasured at the 2012--2013 repeat assessment.

\subsection{Case-Finding Models}\label{sec:case_finding}

All models use one representation of the records before enrollment, fixed in advance and identical for every label, so that it does not favor any of them. Of its 627 columns, 67 describe demographics, GP data supplier, assessment center, calendar time, and registration length, and 48 summarize healthcare use over windows of 30 to 365 days. The other 512 columns, which we call the code history, count primary-care and hospital codes in the year before enrollment, mapped by a fixed hash function to 256 columns per source \citep{Weinberger2009}, which gives the same columns in every fold without learning a vocabulary. Different codes can share a column. Codes for the laboratory tests measured at UK Biobank enrollment, and for their panels, are excluded from both the code history and the healthcare-use counts, and the enrollment assays are never inputs.

We fit $\ell_2$-regularized logistic regression and histogram gradient boosting (scikit-learn; Appendix~\ref{app:models}) in five fixed participant-level folds, for each of the four training labels, with the incident label at three horizons (Table~\ref{tab:terms}). The recorded label predicts $R$ and the \comblabel{} label predicts $R$ or $D$, both in all training participants, so comparing them changes the label but not the population. The measured label predicts $D$ among training participants with $R=0$, the population in which the model is used. The incident label predicts recognition within 1, 2, or 5 years among participants with $R=0$, with inverse-probability-of-censoring weights. Because the measured label needs enrollment assays, we use it as a benchmark for what the recorded label costs.

All models are scored in held-out participants with $R=0$, against $D$. The main measure is the case-finding yield. We also report the yield when 1--20\% are tested, the area under the receiver operating characteristic curve (AUROC), and average precision. To separate prediction of later recognition from detection of a disease-range value at enrollment, we also score each model against recognition within five years, weighting for censoring. Paired 95\% confidence intervals come from 1,000 bootstrap samples within folds, with the top 10\% reselected in each.

\subsection{Recording Propensity}\label{sec:pu_methods}

We estimate the recording propensity with calibrated logistic regression (sigmoid calibration in three inner folds) on the 115 columns without code history, which form $X$ (for example age, sex, GP data supplier, and primary-care contacts and prescriptions in the past year), and evaluate it in held-out assay-positive participants. Standard PU learning assumes a constant propensity \citep{ElkanNoto2008}. Methods that relax this assumption must restrict how the propensity depends on $X$, because recorded data alone cannot separate it from disease risk \citep{BekkerDavis2018,Bekker2020,Gerych2022}. Because $D$ is measured in everyone, our design estimates the propensity without such restrictions. We also test a PU correction that rescales a recorded-label model with the PU propensity $c(X)$, the probability of a prior record among participants who are recorded or assay-positive, using the weight of \citet{ElkanNoto2008} with their constant replaced by $c(X)$ (Appendix~\ref{app:pu}).

\section{Results}

\subsection{The Record Misses Many Disease-Range Values and Catches Up Slowly}\label{sec:recognition}

At enrollment, 19\% of participants with diabetes-range HbA1c, 32\% with CKD-range eGFR, and 62\% with anemia-range hemoglobin had no prior record (Table~\ref{tab:burden}). Among these unrecorded participants, 81\%, 55\%, and 31\% were recognized within five years (Figure~\ref{fig:recognition}A)\@. These estimates barely changed when follow-up was extended to the end of hospital records (Appendix Table~\ref{tab:flow}).

\begin{table}[t]
\floatconts{tab:burden}{\caption{Disease-range values at enrollment and recognition among unrecorded participants. Recognition is cumulative incidence with death as a competing event, with 95\% bootstrap confidence intervals.}}{\centering\footnotesize\setlength{\tabcolsep}{1.6pt}\begin{tabular}{@{}lrrrr@{}}
\toprule
Condition & Assay-positive & Unrecorded & 1 y & 5 y [95\% CI] \\
\midrule
Diabetes range & 6,663 & 1,245 (19\%) & 34\% & 81 [79, 83]\% \\
CKD range & 1,967 & 628 (32\%) & 19\% & 55 [51, 59]\% \\
Anemia range & 7,600 & 4,694 (62\%) & 12\% & 31 [30, 32]\% \\
\bottomrule
\end{tabular}
}
\end{table}

Most unrecorded participants were retested in primary care during follow-up (85--97\% by condition; Appendix Table~\ref{tab:uncoded}). Among those never recognized, 26\% of diabetes-range, 70\% of CKD-range, and 42\% of anemia-range participants had an abnormal primary-care value (Figure~\ref{fig:recognition}B)\@. After a first abnormal value, 81\% of diabetes-range participants were recognized within a year, against 37\% of CKD-range and 38\% of anemia-range participants. After a second abnormal value at least 90 days later, the shares for kidney function and anemia were 36\% and 24\% (Appendix Table~\ref{tab:confirm}). Five-year recognition of anemia-range hemoglobin rose with severity, from 23\% when hemoglobin was less than 0.5 g/dL below the threshold to 54\% when it was 2 g/dL or more below (Appendix Table~\ref{tab:bands}).

\begin{figure*}[t]
\floatconts{fig:recognition}{\caption{\textbf{Recognition and recording.} A, cumulative recognition among unrecorded assay-positive participants, with pointwise bootstrap bands, drawn to six years (participants at risk in Appendix Table~\ref{tab:risk}); the dashed line marks five years. B, follow-up of the 119 diabetes-range, 199 CKD-range, and 2,491 anemia-range participants never recognized by the end of follow-up, in practices in England and Wales. C, among assay-positive participants, the share with a prior record in each decile of the predicted recording propensity; the dashed line is the overall share. D, five-year recognition by group, with 95\% confidence intervals. Q1 is the least and Q5 the most deprived Townsend quintile, and Tested is prior-year testing. Groups with fewer than 30 participants are omitted.}}{\centering\includegraphics[width=\textwidth]{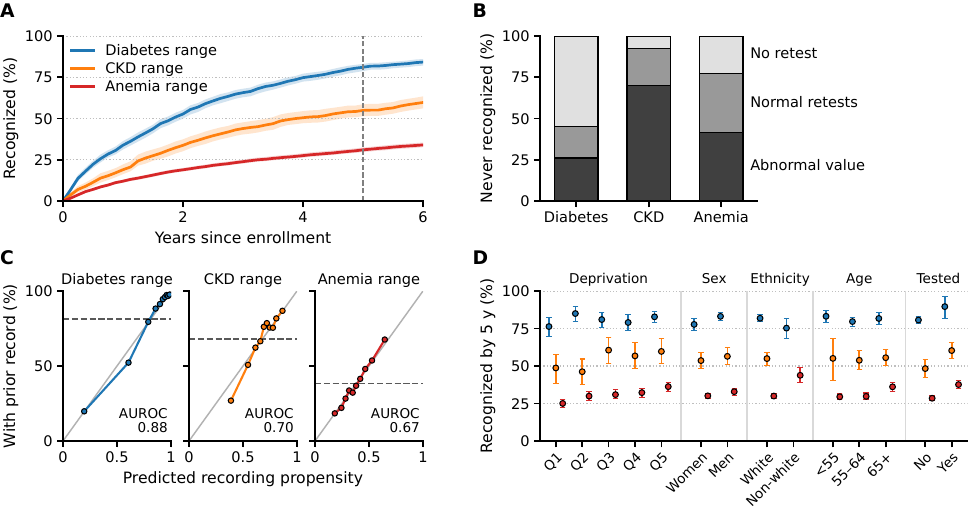}}
\end{figure*}

\subsection{Recording Is Selective}\label{sec:selective}

If recording did not depend on the pre-enrollment features, every decile of the predicted recording propensity would show the same recorded share among assay-positive participants (Figure~\ref{fig:recognition}C, dashed line). Instead, the recorded share of diabetes-range participants rose from 20\% in the lowest decile to 98\% in the highest (AUROC 0.88), and kidney function and anemia showed the same pattern (AUROC 0.70 and 0.67; Appendix Table~\ref{tab:propensity}). Part of this signal may be care that follows a diagnosis. For example, 99\% of diabetes-range participants with an HbA1c test in the year before enrollment already had a record, against 39\% of those without one. Test codes for the enrollment assays are excluded from the features, but primary-care contacts and prescriptions are included. A model trained on recorded diagnoses can learn this signal, and recorded cases are therefore not the random sample of positives that standard PU learning assumes. The PU propensity $c(X)$ was also predictable (AUROC 0.63--0.84; Appendix Table~\ref{tab:propensity}).

Recognition after enrollment also differed between groups (Figure~\ref{fig:recognition}D)\@. For anemia, five-year recognition was 11 percentage points higher in the most than in the least deprived quintile and 14 points higher in non-white than in white participants. It was also higher in older participants and after prior-year testing, and all four differences remained significant after Holm adjustment (Appendix Table~\ref{tab:strata}). For CKD-range eGFR, only prior-year testing differed. For diabetes, no difference was significant after Holm adjustment, although Fine--Gray models adjusted for the other characteristics showed faster recognition with deprivation and prior-year testing and slower recognition in non-white participants (Appendix Table~\ref{tab:severity}).

Severity did not explain the anemia differences. Adding it to the Fine--Gray models left the hazard ratios for deprivation, ethnicity, and prior-year testing almost unchanged (Appendix Tables~\ref{tab:severity} and~\ref{tab:severitygroups}). Removing hemoglobinopathy codes, which are more common in non-white participants, reduced the ethnicity difference only from 14 to 11 points (Appendix Table~\ref{tab:hbp}). The most deprived and the non-white anemia-range participants were also more often recorded at enrollment (45\% and 57\%) than the least deprived and the white participants (34\% and 36\%; Appendix Table~\ref{tab:recordinggroups}).

\begin{figure*}[t]
\floatconts{fig:casefinding}{\caption{\textbf{Case-finding yield by training label.} Models use identical features and are evaluated against the enrollment assays in the same held-out unrecorded participants. A, assay-positive participants found per 1,000 tested when the highest-scoring 10\% are tested, for gradient boosting (filled) and logistic regression (open), with 95\% bootstrap confidence intervals. B, gradient boosting at 1--20\% tested; shading marks the gap between the recorded and measured labels, and the dashed line shows random testing. Incident is the five-year incident label.}}{\centering\includegraphics[width=\textwidth]{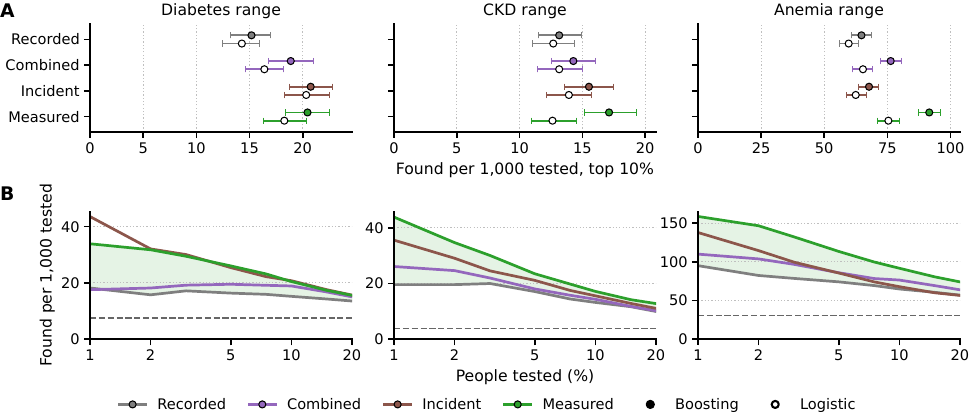}}
\end{figure*}

\subsection{New Diagnoses Often Mark Abnormalities Present at Enrollment}\label{sec:incident}

Among participants first recorded as having diabetes in the year after enrollment, 48\% already had diabetes-range HbA1c at enrollment, and a further 26\% had HbA1c of 42--47.9 mmol/mol. Among those first recorded two to five years after enrollment, 13\% had diabetes-range HbA1c at enrollment (Appendix Table~\ref{tab:composition}). An incident code can therefore mark late recognition of a disease-range value that was present when the prediction was made. A single enrollment value may also not persist. Among participants still unrecorded at the repeat assessment, 43\% of diabetes-range ($n=23$), 70\% of CKD-range ($n=61$), and 26\% of anemia-range ($n=452$) participants were still in the disease range, using creatinine eGFR for kidney function (Appendix Table~\ref{tab:repeat}). We therefore also scored the models against stricter thresholds, which are less affected by within-person variation (Section~\ref{sec:models}).

\subsection{Training on Recorded Diagnoses Finds Fewer Cases}\label{sec:models}
\begin{figure*}[t]
\floatconts{fig:groups}{\caption{\textbf{Recall by group.} Recall is the share of each group's unrecorded assay-positive participants who are among the same overall 10\% tested, for gradient boosting with the recorded and measured labels, with 95\% bootstrap confidence intervals. Asterisks mark differences whose 95\% interval excludes zero (not adjusted for multiplicity). Groups with fewer than 30 assay-positive participants are omitted and marked n $<$ 30.}}{\centering\includegraphics[width=\textwidth]{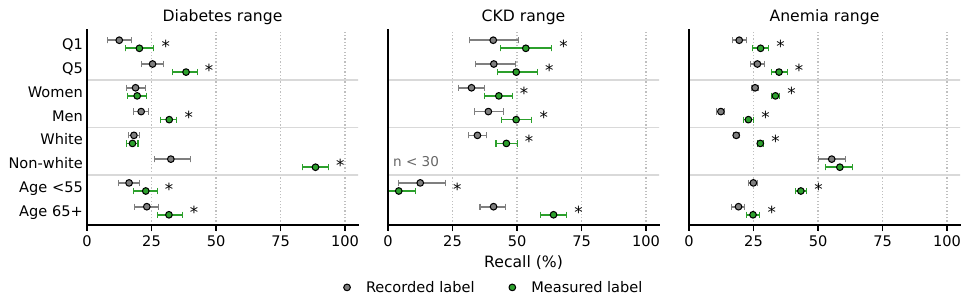}}
\end{figure*}

When the highest-scoring 10\% were tested, gradient boosting trained on the recorded label found 26\%, 23\%, and 29\% fewer assay-positive participants than gradient boosting trained on the measured label, for diabetes, kidney function, and anemia, and all three paired confidence intervals excluded zero (Figure~\ref{fig:casefinding}; Appendix Table~\ref{tab:contrasts}). For anemia, the measured-label model needed 11 tests per case found, against 15 for the recorded-label model and 33 for random testing (Appendix Tables~\ref{tab:operating} and~\ref{tab:netbenefit}). The \comblabel{} label, which adds assay-positive participants to the recorded positives, recovered part of this gap, and the recorded label found 8--20\% fewer than the \comblabel{} label.

Logistic regression gave a similar result for diabetes (22\% fewer) and anemia (21\% fewer). For kidney function, the measured-label logistic model fit poorly with the 512 sparse code-history columns (held-out AUROC 0.71, against 0.81 without these columns), probably because only about 500 unrecorded training participants per fold were assay-positive. Without these columns, the recorded label found 17\% fewer.

The boosting gap held across 1--20\% tested, without the code-history columns, with the published HbA1c calibration, and without medication evidence (Appendix Tables~\ref{tab:contrasts} and~\ref{tab:allmodels}). The recorded label also found fewer participants in 11 of the 12 combinations of condition and GP data supplier (Appendix Table~\ref{tab:provider}). The gap was larger when the models were scored against stricter thresholds, 48\% fewer for hemoglobin below 11 g/dL and 32\% fewer for HbA1c of 53 mmol/mol or more (Appendix Table~\ref{tab:strict}). It narrowed when the 48 healthcare-use columns were also removed, mainly because the measured-label model found fewer without them (Appendix Table~\ref{tab:representation}).

\subsection{Who Is Found}\label{sec:groups}
\begin{figure}[t]
\floatconts{fig:targets}{\caption{\textbf{Two scoring targets.} Difference in AUROC between the measured-label and five-year incident-label boosting models, scored against the enrollment value and against censoring-weighted recognition within five years, with paired 95\% bootstrap confidence intervals.}}{\centering\includegraphics[width=\columnwidth]{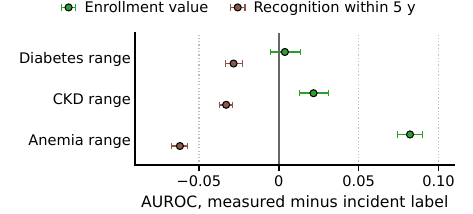}}
\end{figure}

Training on the measured label changed who was selected for testing (Figure~\ref{fig:groups}). For diabetes, the measured-label model selected 74\% of unrecorded non-white participants for testing, against 26\% for the recorded-label model. Recall in this group therefore rose from 32\% to 89\%, while precision among those selected stayed at about 3\% (Appendix Table~\ref{tab:groups}). Recall among white participants did not change, so non-white participants account for the diabetes gap between the two labels. Unrecorded diabetes-range HbA1c was more common in non-white than in white participants (2.6\% against 0.7\%), and part of this difference may reflect higher HbA1c at the same glycemia in South Asian and Black individuals \citep{Mostafa2012,Ziemer2010}. For anemia, recall rose most among participants younger than 55 (from 25\% to 43\%) and nearly doubled among men (from 12\% to 23\%).

\subsection{Later Diagnoses as an Evaluation Target}\label{sec:targets}

The five-year incident label matched the measured label for diabetes, recovered part of the gap for kidney function, and little of it for anemia (Figure~\ref{fig:casefinding}). Figure~\ref{fig:targets} compares the measured-label and five-year incident-label boosting models against both targets (Appendix Table~\ref{tab:targets}). Against anemia-range hemoglobin at enrollment, the measured-label model had an AUROC 0.08 higher than the incident-label model, and against recognition within five years an AUROC 0.06 lower. Kidney function showed the same reversal with smaller differences. For diabetes, the two models were similar against the enrollment value, and the incident-label model was better against later recognition. Among the boosting models for the four training labels, the incident-label model ranked first against later anemia diagnoses and last against anemia-range hemoglobin at enrollment (Appendix Table~\ref{tab:allmodels}). Later diagnoses are therefore a reasonable stand-in for the enrollment value for diabetes, which is recognized quickly and almost completely, but less so for anemia.

\subsection{Propensity Correction}\label{sec:pu}

On the 115 columns without code history, rescaling a calibrated recorded-label model with the estimated PU propensity $c(X)$ found more cases for kidney function and anemia, and also for diabetes with logistic regression. Direct training on the measured label was better than the corrected models for diabetes and for anemia with boosting, and the two did not differ otherwise (Appendix Table~\ref{tab:pucontrasts}). With assays for a random 10\% of training participants, direct training and the correction did not differ for diabetes and kidney function, and direct training was better for anemia (Appendix Tables~\ref{tab:pucontrasts} and~\ref{tab:pu}).

\section{Discussion}

Recorded diagnoses missed many disease-range values at enrollment and caught up slowly for kidney function and anemia. Recording was also selective. It could be predicted from the features a model uses, and for anemia it differed by deprivation and ethnicity in ways that severity did not explain. In our experiments, gradient-boosting models trained on recorded diagnoses found about a quarter fewer unrecorded cases than models trained on the enrollment assays, and evaluation against later diagnoses favored a different anemia model.

A missing record can reflect incomplete capture or a clinical judgment that a borderline value did not need a diagnosis, and our data cannot fully separate the two. For anemia, recognition rose with severity, as judgment about borderline values would predict, but it remained uncommon after repeated abnormal primary-care values. For kidney function and anemia, recognition was also more frequent after prior-year testing, so recording may partly follow contact with care. The most deprived and the non-white anemia-range participants were more often recorded at enrollment and, among those unrecorded, more often recognized within five years. Our data cannot show why these groups differed, or whether they were recognized earlier in the course of disease. For training and evaluating models, these distinctions matter less, because a model trained on recorded diagnoses learns from the record whatever the reason a value went unrecorded.

Similar gaps have been reported elsewhere. In England and Wales, 30\% of people meeting biochemical criteria for stage 3--5 CKD had no CKD code \citep{Kim2018}, and in five other countries 62--96\% of people with stage 3 CKD on two eGFR measurements had no recorded diagnosis \citep{Tangri2023}. The authors of one model that learns from long sequences of diagnosis codes note that it also learns patterns that reflect only the data-collection process \citep{Shmatko2025}. Similarly, a sepsis model performed much worse when evaluated before treatment began, and part of its accuracy therefore reflected clinicians' recognition of sepsis \citep{Kamran2024}.

UK Biobank participants are healthier and less deprived than the general population \citep{Fry2017}, and our models are cross-validated within one biobank. A single assay does not establish chronicity (Section~\ref{sec:incident}), although stricter thresholds gave larger gaps between the recorded and measured labels. Few anemia-range participants were pregnant or unsure whether they were pregnant (19 of 7,600). Unrecorded anemia-range values were mostly normocytic, and microcytic anemia, typical of iron deficiency, was less often unrecorded than normocytic anemia (48\% against 65\%; Appendix Table~\ref{tab:anemiagroups}). Anemia detected outside the health service, for example at blood donation, cannot be seen in these data, and iron prescriptions are an imperfect marker of anemia recognition. Finally, we cannot estimate what finding these cases would achieve, because that depends on treatment effects that our data do not identify.

For people building EHR models to find patients whose condition is not yet recorded, recorded diagnoses are a weak training label on their own. Where a measurement independent of clinical testing exists for some patients, as in research cohorts or screening programs, its positives can be added to the label, or the measurement itself can serve as the training label. Such a measurement can also show whether a PU propensity varies with the features. Models should be evaluated against the condition they are meant to find, because later diagnoses also measure how quickly a condition is recorded.

\acks{This research was conducted using the UK Biobank Resource under application 33127. We thank the participants of UK Biobank for making this work possible. We used AI-based assistants to help with coding and editing and take full responsibility for the paper.}

\clearpage
\bibliography{references}

@techreport{WHOHbA1c2011,
  author      = {{World Health Organization}},
  title       = {Use of glycated haemoglobin ({HbA1c}) in diagnosis of diabetes mellitus: abbreviated report of a {WHO} consultation},
  institution = {World Health Organization},
  address     = {Geneva},
  year        = {2011},
  number      = {WHO/NMH/CHP/CPM/11.1}
}

@techreport{WHOAnaemia2011,
  author      = {{World Health Organization}},
  title       = {Haemoglobin concentrations for the diagnosis of anaemia and assessment of severity},
  institution = {World Health Organization},
  address     = {Geneva},
  year        = {2011},
  number      = {WHO/NMH/NHD/MNM/11.1},
  url         = {https://www.who.int/publications/i/item/WHO-NMH-NHD-MNM-11.1}
}

@article{KDIGO2024,
  author  = {{Kidney Disease: Improving Global Outcomes (KDIGO) CKD Work Group}},
  title   = {{KDIGO} 2024 Clinical Practice Guideline for the Evaluation and Management of Chronic Kidney Disease},
  journal = {Kidney International},
  year    = {2024},
  volume  = {105},
  number  = {4S},
  pages   = {S117--S314},
  doi     = {10.1016/j.kint.2023.10.018}
}

@article{Young2022,
  author  = {Young, Katherine G. and McGovern, Andrew P. and Barroso, In{\^e}s and Hattersley, Andrew T. and Jones, Angus G. and Shields, Beverley M. and Thomas, Nicholas J. and Dennis, John M.},
  title   = {The impact of population-level {HbA1c} screening on reducing diabetes diagnostic delay in middle-aged adults: a {UK} {Biobank} analysis},
  journal = {Diabetologia},
  year    = {2023},
  volume  = {66},
  number  = {2},
  pages   = {300--309},
  doi     = {10.1007/s00125-022-05824-0}
}

@article{Sisk2024,
  author  = {Sisk, Rose and Cameron, Rory and Tahir, Waqas and Sammut-Powell, Camilla},
  title   = {Diagnosis codes underestimate chronic kidney disease incidence compared with {eGFR}-based evidence: a retrospective observational study of patients with type 2 diabetes in {UK} primary care},
  journal = {BJGP Open},
  year    = {2024},
  volume  = {8},
  number  = {1},
  pages   = {BJGPO.2023.0079},
  doi     = {10.3399/BJGPO.2023.0079}
}

@article{Schulz2021,
  author  = {Schulz, Wade L. and Young, H. Patrick and Coppi, Andreas and Mortazavi, Bobak J. and Lin, Zhenqiu and Jean, Raymond A. and Krumholz, Harlan M.},
  title   = {Temporal relationship of computed and structured diagnoses in electronic health record data},
  journal = {BMC Medical Informatics and Decision Making},
  year    = {2021},
  volume  = {21},
  number  = {1},
  pages   = {61},
  doi     = {10.1186/s12911-021-01416-x}
}

@inproceedings{BekkerDavis2018,
  author    = {Bekker, Jessa and Davis, Jesse},
  title     = {Learning from positive and unlabeled data under the selected at random assumption},
  booktitle = {Proceedings of the Second International Workshop on Learning with Imbalanced Domains: Theory and Applications},
  series    = {PMLR},
  volume    = {94},
  pages     = {8--22},
  year      = {2018}
}

@inproceedings{ElkanNoto2008,
  author    = {Elkan, Charles and Noto, Keith},
  title     = {Learning classifiers from only positive and unlabeled data},
  booktitle = {Proceedings of the 14th ACM SIGKDD International Conference on Knowledge Discovery and Data Mining},
  pages     = {213--220},
  year      = {2008},
  doi       = {10.1145/1401890.1401920}
}

@inproceedings{Chang2022,
  author    = {Chang, Trenton and Sjoding, Michael W. and Wiens, Jenna},
  title     = {Disparate censorship \& undertesting: a source of label bias in clinical machine learning},
  booktitle = {Proceedings of the 7th Machine Learning for Healthcare Conference},
  series    = {PMLR},
  volume    = {182},
  pages     = {343--390},
  year      = {2022}
}

@inproceedings{ChangWiens2024,
  author    = {Chang, Trenton and Wiens, Jenna},
  title     = {From biased selective labels to pseudo-labels: an expectation-maximization framework for learning from biased decisions},
  booktitle = {Proceedings of the 41st International Conference on Machine Learning},
  series    = {PMLR},
  volume    = {235},
  pages     = {6286--6324},
  year      = {2024}
}

@article{Obermeyer2019,
  author  = {Obermeyer, Ziad and Powers, Brian and Vogeli, Christine and Mullainathan, Sendhil},
  title   = {Dissecting racial bias in an algorithm used to manage the health of populations},
  journal = {Science},
  year    = {2019},
  volume  = {366},
  number  = {6464},
  pages   = {447--453},
  doi     = {10.1126/science.aax2342}
}

@article{ZangerTishler2024,
  author  = {Zanger-Tishler, Michael and Nyarko, Julian and Goel, Sharad},
  title   = {Risk scores, label bias, and everything but the kitchen sink},
  journal = {Science Advances},
  year    = {2024},
  volume  = {10},
  number  = {13},
  pages   = {eadi8411},
  doi     = {10.1126/sciadv.adi8411}
}

@article{Agniel2018,
  author  = {Agniel, Denis and Kohane, Isaac S. and Weber, Griffin M.},
  title   = {Biases in electronic health record data due to processes within the healthcare system: retrospective observational study},
  journal = {BMJ},
  year    = {2018},
  volume  = {361},
  pages   = {k1479},
  doi     = {10.1136/bmj.k1479}
}

@article{Sudlow2015,
  author  = {Sudlow, Cathie and Gallacher, John and Allen, Naomi and Beral, Valerie and Burton, Paul and Danesh, John and Downey, Paul and Elliott, Paul and Green, Jane and Landray, Martin and Liu, Bette and Matthews, Paul and Ong, Giok and Pell, Jill and Silman, Alan and Young, Alan and Sprosen, Tim and Peakman, Tim and Collins, Rory},
  title   = {{UK} {Biobank}: an open access resource for identifying the causes of a wide range of complex diseases of middle and old age},
  journal = {PLoS Medicine},
  year    = {2015},
  volume  = {12},
  number  = {3},
  pages   = {e1001779},
  doi     = {10.1371/journal.pmed.1001779}
}

@article{Inker2021,
  author  = {Inker, Lesley A. and Eneanya, Nwamaka D. and Coresh, Josef and others},
  title   = {New creatinine- and cystatin {C}-based equations to estimate {GFR} without race},
  journal = {New England Journal of Medicine},
  year    = {2021},
  volume  = {385},
  number  = {19},
  pages   = {1737--1749},
  doi     = {10.1056/NEJMoa2102953}
}

@article{Shmatko2025,
  author  = {Shmatko, Artem and Jung, Alexander Wolfgang and Gaurav, Kumar and Brunak, S{\o}ren and Mortensen, Laust Hvas and Birney, Ewan and Fitzgerald, Tom and Gerstung, Moritz},
  title   = {Learning the natural history of human disease with generative transformers},
  journal = {Nature},
  year    = {2025},
  volume  = {647},
  number  = {8088},
  pages   = {248--256},
  doi     = {10.1038/s41586-025-09529-3}
}

@article{Fry2017,
  author  = {Fry, Anna and Littlejohns, Thomas J. and Sudlow, Cathie and Doherty, Nicola and Adamska, Ligia and Sprosen, Tim and Collins, Rory and Allen, Naomi E.},
  title   = {Comparison of sociodemographic and health-related characteristics of {UK} {Biobank} participants with those of the general population},
  journal = {American Journal of Epidemiology},
  year    = {2017},
  volume  = {186},
  number  = {9},
  pages   = {1026--1034},
  doi     = {10.1093/aje/kwx246}
}

@article{Aalen1978,
  author  = {Aalen, Odd O. and Johansen, S{\o}ren},
  title   = {An empirical transition matrix for non-homogeneous {Markov} chains based on censored observations},
  journal = {Scandinavian Journal of Statistics},
  year    = {1978},
  volume  = {5},
  number  = {3},
  pages   = {141--150}
}

@article{YoungCalibration2022,
  author = {Young, Katherine G. and McDonald, Timothy J. and Shields, Beverley M.},
  title = {Glycated haemoglobin measurements from {UK} {Biobank} are different to those in linked primary care records: implications for combining biochemistry data from research studies and routine clinical care},
  journal = {International Journal of Epidemiology},
  year = {2022}, volume = {51}, number = {3}, pages = {1022--1024},
  doi = {10.1093/ije/dyab265}
}

@article{He2021,
  author = {He, Yixuan and Lakhani, Chirag M. and Rasooly, Danielle and Manrai, Arjun K. and Tzoulaki, Ioanna and Patel, Chirag J.},
  title = {Comparisons of Polyexposure, Polygenic, and Clinical Risk Scores in Risk Prediction of Type 2 Diabetes},
  journal = {Diabetes Care}, year = {2021}, volume = {44}, number = {4}, pages = {935--943},
  doi = {10.2337/dc20-2049}
}

@article{Rodwell2026,
  author = {Rodwell, Graham and Ioannidis, John P. A. and Kim, Stuart K.},
  title = {A genetic and clinical risk factor algorithm to aid in identifying new cases of chronic kidney disease from the general population},
  journal = {Frontiers in Genetics}, year = {2026}, volume = {17}, pages = {1799312},
  doi = {10.3389/fgene.2026.1799312}
}

@article{DiCai2026,
  author = {Di, Yazheng and Cai, Na},
  title = {Learning lifetime disease liability reveals and removes genetic confounding in electronic health records},
  journal = {medRxiv}, year = {2026}, note = {Preprint},
  doi = {10.64898/2026.02.15.26346336}
}

@article{Garg2024,
  author = {Garg, Manik and Karpinski, Marcin and Matelska, Dorota and Middleton, Lawrence and Burren, Oliver S. and Hu, Fengyuan and Wheeler, Eleanor and Smith, Katherine R. and Fabre, Margarete A. and Mitchell, Jonathan and O'Neill, Amanda and Ashley, Euan A. and Harper, Andrew R. and Wang, Quanli and Dhindsa, Ryan S. and Petrovski, Slav{\'e} and Vitsios, Dimitrios},
  title = {Disease prediction with multi-omics and biomarkers empowers case--control genetic discoveries in the {UK} {Biobank}},
  journal = {Nature Genetics}, year = {2024}, volume = {56}, number = {9}, pages = {1821--1831},
  doi = {10.1038/s41588-024-01898-1}
}

@article{Inker2012,
  author = {Inker, Lesley A. and Schmid, Christopher H. and Tighiouart, Hocine and Eckfeldt, John H. and Feldman, Harold I. and Greene, Tom and Kusek, John W. and Manzi, Jane and Van Lente, Frederick and Zhang, Yaping Lucy and Coresh, Josef and Levey, Andrew S.},
  title = {Estimating glomerular filtration rate from serum creatinine and cystatin {C}},
  journal = {New England Journal of Medicine}, year = {2012}, volume = {367}, number = {1}, pages = {20--29},
  doi = {10.1056/NEJMoa1114248}
}

@article{FineGray1999,
  author  = {Fine, Jason P. and Gray, Robert J.},
  title   = {A proportional hazards model for the subdistribution of a competing risk},
  journal = {Journal of the American Statistical Association},
  year    = {1999},
  volume  = {94},
  number  = {446},
  pages   = {496--509},
  doi     = {10.1080/01621459.1999.10474144}
}

@misc{BenitezAurioles2026,
  author        = {Benitez-Aurioles, Jose and Silva, Ricardo and McMillan, Brian and Sperrin, Matthew},
  title         = {Correcting heterogeneous diagnostic bias when developing clinical prediction models using causal hidden {Markov} models},
  year          = {2026},
  eprint        = {2605.06059},
  archivePrefix = {arXiv},
  primaryClass  = {stat.AP},
  howpublished  = {arXiv:2605.06059}
}

@article{Kim2018,
  author  = {Kim, Lois G. and Cleary, Faye and Wheeler, David C. and Caplin, Ben and Nitsch, Dorothea and Hull, Sally A.},
  title   = {How do primary care doctors in {England} and {Wales} code and manage people with chronic kidney disease? {Results} from the {National Chronic Kidney Disease Audit}},
  journal = {Nephrology Dialysis Transplantation},
  year    = {2018},
  volume  = {33},
  number  = {8},
  pages   = {1373--1379},
  doi     = {10.1093/ndt/gfx280}
}

@article{Tangri2023,
  author  = {Tangri, Navdeep and Moriyama, Toshiki and Schneider, Markus P. and Virgitti, Jean Blaise and De Nicola, Luca and Arnold, Matthew and Barone, Salvatore and Peach, Emily and Wittbrodt, Eric and Chen, Hungta and J{\"a}rbrink, Krister and Kushner, Pamela},
  title   = {Prevalence of undiagnosed stage 3 chronic kidney disease in {France}, {Germany}, {Italy}, {Japan} and the {USA}: results from the multinational observational {REVEAL-CKD} study},
  journal = {BMJ Open},
  year    = {2023},
  volume  = {13},
  number  = {5},
  pages   = {e067386},
  doi     = {10.1136/bmjopen-2022-067386}
}

@inproceedings{Weinberger2009,
  author    = {Weinberger, Kilian and Dasgupta, Anirban and Langford, John and Smola, Alex and Attenberg, Josh},
  title     = {Feature hashing for large scale multitask learning},
  booktitle = {Proceedings of the 26th Annual International Conference on Machine Learning},
  pages     = {1113--1120},
  year      = {2009},
  doi       = {10.1145/1553374.1553516}
}

@inproceedings{Bekker2020,
  author    = {Bekker, Jessa and Robberechts, Pieter and Davis, Jesse},
  title     = {Beyond the selected completely at random assumption for learning from positive and unlabeled data},
  booktitle = {Machine Learning and Knowledge Discovery in Databases (ECML PKDD 2019)},
  publisher = {Springer},
  pages     = {71--85},
  year      = {2020},
  doi       = {10.1007/978-3-030-46147-8_5}
}

@article{Gerych2022,
  author  = {Gerych, Walter and Hartvigsen, Thomas and Buquicchio, Luke and Agu, Emmanuel and Rundensteiner, Elke},
  title   = {Recovering the propensity score from biased positive unlabeled data},
  journal = {Proceedings of the AAAI Conference on Artificial Intelligence},
  year    = {2022},
  volume  = {36},
  number  = {6},
  pages   = {6694--6702},
  doi     = {10.1609/aaai.v36i6.20624}
}

@article{Kamran2024,
  author  = {Kamran, Fahad and Tjandra, Donna and Heiler, Andrew and Virzi, Jessica and Singh, Karandeep and King, Jessie E. and Valley, Thomas S. and Wiens, Jenna},
  title   = {Evaluation of sepsis prediction models before onset of treatment},
  journal = {NEJM AI},
  year    = {2024},
  volume  = {1},
  number  = {3},
  doi     = {10.1056/AIoa2300032}
}

@inproceedings{Balachandar2024,
  author    = {Balachandar, Sidhika and Garg, Nikhil and Pierson, Emma},
  title     = {Domain constraints improve risk prediction when outcome data is missing},
  booktitle = {The Twelfth International Conference on Learning Representations (ICLR)},
  year      = {2024}
}

@article{Mostafa2012,
  author  = {Mostafa, Samiul A. and Davies, Melanie J. and Webb, David R. and Srinivasan, Balasubramanian Thiagarajan and Gray, Laura J. and Khunti, Kamlesh},
  title   = {Independent effect of ethnicity on glycemia in {South} {Asians} and white {Europeans}},
  journal = {Diabetes Care},
  year    = {2012},
  volume  = {35},
  number  = {8},
  pages   = {1746--1748},
  doi     = {10.2337/dc11-2079}
}

@article{Ziemer2010,
  author  = {Ziemer, David C. and Kolm, Paul and Weintraub, William S. and Vaccarino, Viola and Rhee, Mary K. and Twombly, Jennifer G. and Narayan, K. M. Venkat and Koch, David D. and Phillips, Lawrence S.},
  title   = {Glucose-independent, black--white differences in hemoglobin {A1c} levels: a cross-sectional analysis of 2 studies},
  journal = {Annals of Internal Medicine},
  year    = {2010},
  volume  = {152},
  number  = {12},
  pages   = {770--777},
  doi     = {10.7326/0003-4819-152-12-201006150-00004}
}
\appendix
\onecolumn
\section{Definitions, Code Lists, and Data Sources}\label{app:defs}

\paragraph{Data Sources.} Primary-care records in UK Biobank come from four GP data suppliers, TPP (England, coded in Clinical Terms Version 3, CTV3), Vision (England, Read Version 2), and the Scottish and Welsh extracts (Read Version 2). Hospital records are Hospital Episode Statistics (England) and their Scottish and Welsh equivalents, coded in ICD-10.

\paragraph{Enrollment Assays ($D$).} HbA1c is UK Biobank field 30750, with a threshold of 48 mmol/mol (the World Health Organization and National Institute for Health and Care Excellence diagnostic threshold); stricter analyses use 53 mmol/mol. Creatinine eGFR uses the race-free 2021 CKD-EPI equation with field 30700, age, and sex, and cystatin C eGFR uses the 2012 CKD-EPI cystatin C equation with field 30720. The CKD range requires both to be below 60 mL/min/1.73 m$^2$; stricter analyses require both to be below 45. Hemoglobin is field 30020, with thresholds of 12.0 g/dL in women and 13.0 g/dL in men; stricter analyses use 11.0 g/dL\@. All are single enrollment measurements.

\paragraph{Prior Record ($R$).} Self-reported illness codes (field 20002: diabetes 1220, 1222, 1223; kidney 1192, 1193, 1194, 1519, 1520, 1607, 1608, 1609; anemia 1446, 1330, 1331, 1332), doctor-diagnosed diabetes (field 2443), insulin use (fields 6153, 6177), self-reported medication (field 20003, matched by name to the drug lists below), any UK Biobank first-occurrence date on or before enrollment (fields 130706--130714 for ICD-10 E10--E14, 132032 for N18, and 130622--130648 for D50--D64), primary-care diagnosis codes in the lists below, and primary-care prescriptions in the drug lists on or before enrollment. The ICD-10 range D50--D64 includes the hemoglobinopathies (D56--D58), so thalassemia and sickle-cell codes count as anemia records; Table~\ref{tab:hbp} removes them.

\paragraph{Recognition.} The earliest of a first-occurrence date whose source (Data-Coding 2171) is primary care (30, 31), hospital (40, 41), or death (20, 21), a primary-care diagnosis code from the lists, and a primary-care prescription in the drug lists after enrollment. Primary-care diagnosis lists are the Read 2 and CTV3 codes that UK Biobank's lookup files map to ICD-10 E10--E14 (diabetes), D50--D64 (anemia), or N18 (CKD)\@. For CKD we add all Read 2 codes under 1Z1 (chronic kidney disease stages), which the UK Biobank Read-2-to-ICD-10 map does not include, and Read 2 and CTV3 codes whose term names chronic kidney disease, chronic renal failure, impairment, or insufficiency, or end-stage renal disease, excluding terms about risk, monitoring, screening, review, family history, administration, education, and acute disease. The lists contain 60 Read 2 and 144 CTV3 codes for CKD, 221 and 280 for diabetes, and 208 and 353 for anemia.

\paragraph{Drug Lists.} Prescriptions are matched by drug name, using the name on the prescription record or, for Read 2 drug codes, the term in the UK Biobank lookup. The diabetes list contains glucose-lowering drugs (insulins, metformin, sulfonylureas, meglitinides, thiazolidinediones, DPP-4 and SGLT2 inhibitors, GLP-1 receptor agonists, and acarbose, with their brand names), and excludes monitoring and injection devices and GLP-1 receptor agonists at weight-management doses. The anemia list contains oral and parenteral iron and excludes iron combined with folic acid, multivitamins with iron, folic acid or vitamin B12 alone, and iron chelators. We do not use British National Formulary (BNF) codes, which only two suppliers provide and which group these drugs with related items such as diabetes monitoring supplies, folic acid, and vitamin B12. The same lists define prior record and recognition.

\paragraph{Primary-Care Values.} HbA1c uses Read codes 42W5., 42W4., and 42W..\ and CTV3 codes XaPbt and XaERp; values recorded as a percentage (DCCT) are converted to mmol/mol and rounded to an integer, so 6.5\% becomes 48. Creatinine uses 44J3.\ and XE2q5 and is converted to eGFR with the 2021 CKD-EPI equation at the age of the test. Hemoglobin uses 423.., XE2m6, and Xa96v, with values of 40 or more read as g/L\@. In the Scottish extract, numeric results are often stored with an operator in a separate field, and part of them cannot be decoded, so analyses of primary-care values after enrollment use practices in England and Wales. Prior-year testing uses the presence of a test code, not its value, and is available for all suppliers.

\paragraph{Follow-Up.} Recognition analyses end at the earlier of the participant's primary-care extract end and the hospital-record end (31 October 2022 for England, 31 July 2021 for Scotland, and 28 February 2018 for Wales); primary-care extracts end in 2016--2017. Death competes with recognition, and a diagnosis on a death record counts as recognition on the date of death. Estimates among all UK Biobank participants run to the end of hospital records, so their recognition rate also reflects missing primary-care records. The curves in Figure~\ref{fig:recognition}A show pointwise bands from 300 bootstrap samples and are drawn to six years, when 172 diabetes-range and 213 CKD-range participants remain at risk (Table~\ref{tab:risk}). Reported intervals at fixed horizons use 1,000 samples, except the hospital-end rows of Table~\ref{tab:flow}, which use 300.

\paragraph{Group Comparisons.} Groups are sex, Townsend deprivation quintile, self-reported ethnicity (white versus Asian, Black, mixed, Chinese, or other), age at enrollment, and prior-year testing. Comparisons use paired participant-bootstrap differences in five-year Aalen--Johansen recognition with 1,000 resamples and Holm adjustment within each condition. Fine--Gray models are fitted on the same participants without and with a three-degree-of-freedom natural spline of severity, and the kidney model includes splines of both eGFR values. Standard errors are clustered by participant.

\paragraph{Definition Sensitivities.} We apply the published calibration $\mathrm{HbA1c}_{\mathrm{cal}}=0.9696\,\mathrm{HbA1c}_{\mathrm{UKB}}+3.3595$, which maps UK Biobank values onto the primary-care scale, before thresholding at 48 mmol/mol \citep{YoungCalibration2022}. The calibration was derived from UK Biobank participants with linked primary-care measurements, so it is not independent of our cohort, and we did not refit it. A second sensitivity analysis removes all prescription and self-reported medication evidence from both the prior record and recognition, keeping diagnoses and self-reported illnesses. We refit all models under each definition (Table~\ref{tab:allmodels}).

\section{Model Implementation}\label{app:models}

\paragraph{Models and Preprocessing.} Logistic regression uses $\ell_2$ regularization with $C=0.1$ and the L-BFGS solver. Gradient boosting (scikit-learn \texttt{HistGradientBoostingClassifier}) uses at most 200 iterations, learning rate 0.05, at most 15 leaves, at least 50 samples per leaf, and $\ell_2$ regularization 1, with the default early stopping, which holds out 10\% of the training data and stops after 10 iterations without improvement in loss. Features are standardized within each outer training fold. The features and outer folds are the same for all labels. Fitting uses only the outer training fold, and no held-out assay, recognition outcome, or performance metric enters model selection. In the code history, each code is assigned to one of the 256 columns of its source (primary care or hospital) by the first eight bytes, taken modulo 256, of the SHA-256 hash of a string that joins a fixed prefix, the source table, the coding system, and the code.

\paragraph{Horizon Outcomes.} Let $G(t)$ be the survival function of administrative censoring. Recognition before death and by horizon $h$ is positive. Death without recognition by $h$ is negative, as is remaining alive and not recognized through $h$. An observed recognition or death at $t\leq h$ receives weight $1/\hat G(t-)$, and follow-up through $h$ receives $1/\hat G(h-)$. Earlier administrative censoring receives zero weight. Censoring models are Kaplan--Meier estimators in unrecorded outer-training participants, stratified by GP data supplier and enrollment year; strata with fewer than 200 training participants use the pooled estimator. Training weights are normalized to mean one. This targets cumulative recognition before death under independent censoring within these strata.

\paragraph{Screening and Uncertainty.} Within each test fold we select the highest-scoring $\operatorname{round}(qn)$ participants, where $q$ is the fraction tested and $n$ the number of participants in the fold, breaking ties by participant identifier. Bootstrap resampling preserves each fold's size, selection is repeated within each resampled population, and the same resamples are used for all models. Intervals condition on the fitted models. The averaged yield is the trapezoidal average of the yield curve over 1--20\% tested, evaluated at 1, 2, 3, 5, 7.5, 10, 15, and 20\%. AUROC accounts for ties and uses censoring weights when scored against recognition. Group recall uses the same overall selection rather than selecting the top 10\% within each group.

\section{Positive--Unlabeled Interpretation and Implementation}\label{app:pu}

Let $T=R\lor D$, positive if a participant is recorded or assay-positive, so that $R\leq T$ by construction. Writing $p(X)=P(T=1\mid X)$, $g(X)=P(R=1\mid X)$, and $c(X)=P(R=1\mid T=1,X)$ gives $g=pc$. The recording propensity in the main text, $P(R=1\mid D=1,X)$, conditions on the enrollment value instead; $c(X)$ is the quantity that PU learning uses. Because case finding is applied where $R=0$, Bayes' rule gives
\[
P(T=1\mid R=0,X)=\frac{P(T=1,R=0\mid X)}{P(R=0\mid X)}=\frac{p-g}{1-g}=\frac{g(1-c)}{c(1-g)}.
\]
This is the weight that \citet{ElkanNoto2008} assign to unlabeled examples, with their constant $c$ replaced by $c(X)$. Within $R=0$, $T=D$, so this is the probability that case finding needs, the probability that an unrecorded participant is assay-positive. The population score $g/c$ answers a different question. When $c$ is constant, both are monotone in $g$; when $c$ depends on $X$, their rankings can differ. Because $T$ includes recorded history, it stands for disease history rather than a disease-range value at enrollment, and treated historical positives can have $R=1$ and $D=0$.

\paragraph{Estimation.} All PU components use the 115-column representation. The $g$ models and the propensity models use sigmoid calibration with three stratified inner folds within each outer training fold, and propensity logistic regression uses $C=1$. We fit $c$ among participants who are recorded or assay-positive, and the recording propensity among assay-positive participants. Both are validated in all held-out participants of the corresponding set, including those already recorded. For numerical stability we clip $g$ to $[10^{-8},1-10^{-8}]$, $c$ to $[0.02,1]$, and $p=g/c$ to $[g,1]$; we also evaluate a floor of 0.005 for $c$. The corrected scores are compared with a calibrated recorded-label model and with measured-label training on the same 115 columns.

\paragraph{Assays for 10\% of Training Participants.} In each outer training fold we randomly select 10\% of participants with the required assay, before inspecting $D$, and use their assay values. Recorded status remains available for the whole training fold. The propensity is fitted among participants in this subset who are recorded or assay-positive, and the $g$ model uses all recorded labels. Direct training uses the measured label, $D$ among unrecorded participants in the same subset. Both use logistic regression on the 115 columns. This is one fixed random allocation per fold, and the intervals condition on it.

\section{Additional Results}\label{app:results}

All tables use the cohort, folds, features, and estimators of the main analysis unless stated otherwise.

\subsection{Recognition and Follow-Up Testing}
{\footnotesize\setlength{\tabcolsep}{2.5pt}
\begin{longtable}{lrrrrrr}
\caption{Assays, disease-range values, and recognition in the study cohort (participants with primary-care records covering the year before enrollment) and among all UK Biobank participants. The rows ``to hospital end'' extend follow-up of the same unrecorded study-cohort participants to the end of hospital records. Recognition among all participants also runs to the end of hospital records and is lower because many lack primary-care records. Creatinine-only CKD range requires only creatinine eGFR below 60; the other added rows are the definition variants in Appendix~\ref{app:defs}.} \label{tab:flow} \\
\toprule
Condition & Cohort & Assayed & Assay-positive & Unrecorded & Followed & 5-y recognition (\%) \\
\midrule
\endfirsthead
\caption[]{Assays, disease-range values, and recognition in the study cohort (participants with primary-care records covering the year before enrollment) and among all UK Biobank participants. The rows ``to hospital end'' extend follow-up of the same unrecorded study-cohort participants to the end of hospital records. Recognition among all participants also runs to the end of hospital records and is lower because many lack primary-care records. Creatinine-only CKD range requires only creatinine eGFR below 60; the other added rows are the definition variants in Appendix~\ref{app:defs}.} \\
\toprule
Condition & Cohort & Assayed & Assay-positive & Unrecorded & Followed & 5-y recognition (\%) \\
\midrule
\endhead
\midrule
\multicolumn{7}{r}{Continued on next page} \\
\midrule
\endfoot
\bottomrule
\endlastfoot
Diabetes range & study cohort & 174,920 & 6,663 & 1,245 & 1,244 & 81.1 [78.8, 83.2] \\
Diabetes range & to hospital end & -- & -- & -- & 1,245 & 80.8 [78.4, 82.6] \\
Diabetes range & all participants & 466,377 & 17,606 & 3,300 & 3,300 & 52.0 [50.3, 53.7] \\
CKD range & study cohort & 175,209 & 1,967 & 628 & 628 & 54.9 [51.2, 58.8] \\
CKD range & to hospital end & -- & -- & -- & 628 & 54.5 [51.0, 58.1] \\
CKD range & all participants & 468,884 & 5,036 & 3,144 & 3,144 & 28.1 [26.5, 29.7] \\
Anemia range & study cohort & 178,163 & 7,600 & 4,694 & 4,686 & 31.0 [29.7, 32.4] \\
Anemia range & to hospital end & -- & -- & -- & 4,694 & 30.8 [29.4, 32.1] \\
Anemia range & all participants & 478,042 & 20,589 & 15,537 & 15,537 & 19.4 [18.8, 20.0] \\
Creatinine-only CKD range & study cohort & 175,351 & 2,898 & 1,346 & 1,344 & 35.1 [32.6, 37.4] \\
Calibrated HbA1c & study cohort & 174,920 & 7,928 & 1,890 & 1,889 & 70.9 [68.9, 72.8] \\
Diabetes, no medication & study cohort & 174,920 & 6,663 & 1,251 & 1,250 & 81.2 [79.0, 83.5] \\*
Anemia, no medication & study cohort & 178,163 & 7,600 & 5,532 & 5,523 & 22.8 [21.7, 23.9] \\
\end{longtable}
}

{\small\setlength{\tabcolsep}{3.5pt}
\begin{longtable}{lrrrrr}
\caption{Primary-care testing after enrollment among unrecorded assay-positive participants in practices in England and Wales. Retested and abnormal-value shares use the Followed count as denominator. The last two columns start a new clock at the first abnormal primary-care value, exclude people recognized before it, and count same-day recognition. For kidney function, an abnormal value is creatinine eGFR below 60.} \label{tab:uncoded} \\
\toprule
Condition & Followed & Retested & Abnormal value & At abnormal value & 1-y recognition (\%) \\
\midrule
\endfirsthead
\caption[]{Primary-care testing after enrollment among unrecorded assay-positive participants in practices in England and Wales. Retested and abnormal-value shares use the Followed count as denominator. The last two columns start a new clock at the first abnormal primary-care value, exclude people recognized before it, and count same-day recognition. For kidney function, an abnormal value is creatinine eGFR below 60.} \\
\toprule
Condition & Followed & Retested & Abnormal value & At abnormal value & 1-y recognition (\%) \\
\midrule
\endhead
\midrule
\multicolumn{6}{r}{Continued on next page} \\
\midrule
\endfoot
\bottomrule
\endlastfoot
Diabetes range & 1,088 & 93\% & 91\% & 549 & 81.0 [77.5, 84.6] \\*
CKD range & 529 & 97\% & 88\% & 431 & 37.4 [33.2, 41.9] \\*
Anemia range & 4,062 & 85\% & 60\% & 2,262 & 38.1 [36.0, 40.2] \\*
Creatinine-only CKD range & 1,187 & 94\% & 63\% & 695 & 34.3 [30.7, 38.0] \\
\end{longtable}
}

{\small\setlength{\tabcolsep}{3.5pt}
\begin{longtable}{lrrrrr}
\caption{Recognition within one year of an abnormal primary-care value among unrecorded assay-positive participants in practices in England and Wales. Same-day recognition counts.} \label{tab:confirm} \\
\toprule
Condition & Clock starts at & With value & Rec.\ before & At risk & 1-y rec.\ (\%) \\
\midrule
\endfirsthead
\caption[]{Recognition within one year of an abnormal primary-care value among unrecorded assay-positive participants in practices in England and Wales. Same-day recognition counts.} \\
\toprule
Condition & Clock starts at & With value & Rec.\ before & At risk & 1-y rec.\ (\%) \\
\midrule
\endhead
\midrule
\multicolumn{6}{r}{Continued on next page} \\
\midrule
\endfoot
\bottomrule
\endlastfoot
Diabetes range & First abnormal value & 985 & 436 & 549 & 81 [78, 85] \\*
Diabetes range & Second abnormal value, $\geq$90 days later & 918 & 835 & 83 & 69 [59, 79] \\*
CKD range & First abnormal value & 464 & 33 & 431 & 37 [33, 42] \\*
CKD range & Second abnormal value, $\geq$90 days later & 409 & 151 & 258 & 36 [30, 42] \\*
Anemia range & First abnormal value & 2,433 & 171 & 2,262 & 38 [36, 40] \\*
Anemia range & Second abnormal value, $\geq$90 days later & 1,566 & 561 & 1,005 & 24 [21, 26] \\
\end{longtable}
}

{\small\setlength{\tabcolsep}{3.5pt}
\begin{longtable}{lrrrrr}
\caption{Prior recording and five-year recognition by severity. Bands with fewer than 30 unrecorded participants are omitted. Hb deficit is the shortfall of hemoglobin below the sex-specific threshold, and creatinine eGFR is in mL/min/1.73 m$^2$.} \label{tab:bands} \\
\toprule
Condition & Scale & Band & Assay-positive & Unrecorded & 5-y rec.\ (\%) [95\% CI] \\
\midrule
\endfirsthead
\caption[]{Prior recording and five-year recognition by severity. Bands with fewer than 30 unrecorded participants are omitted. Hb deficit is the shortfall of hemoglobin below the sex-specific threshold, and creatinine eGFR is in mL/min/1.73 m$^2$.} \\
\toprule
Condition & Scale & Band & Assay-positive & Unrecorded & 5-y rec.\ (\%) [95\% CI] \\
\midrule
\endhead
\midrule
\multicolumn{6}{r}{Continued on next page} \\
\midrule
\endfoot
\bottomrule
\endlastfoot
Diabetes range & HbA1c (mmol/mol) & 48--52.9 & 2,204 & 29\% & 72 [69, 75] \\*
Diabetes range & HbA1c (mmol/mol) & 53--63.9 & 2,542 & 13\% & 91 [87, 94] \\*
Diabetes range & HbA1c (mmol/mol) & $\geq$64 & 1,917 & 14\% & 91 [88, 94] \\*
CKD range & Creatinine eGFR & 30--44.9 & 392 & 19\% & 56 [44, 67] \\*
CKD range & Creatinine eGFR & 45--59.9 & 1,429 & 38\% & 54 [50, 59] \\*
Anemia range & Hb deficit (g/dL) & $<$0.5 & 3,999 & 67\% & 23 [21, 24] \\*
Anemia range & Hb deficit (g/dL) & 0.5--0.99 & 1,814 & 60\% & 36 [33, 39] \\*
Anemia range & Hb deficit (g/dL) & 1--1.99 & 1,237 & 51\% & 47 [43, 51] \\*
Anemia range & Hb deficit (g/dL) & $\geq$2 & 550 & 52\% & 54 [48, 60] \\
\end{longtable}
}

{\small\setlength{\tabcolsep}{3.5pt}
\begin{longtable}{lrrrrr}
\caption{Participants still at risk just before each horizon in the recognition curves of Figure~\ref{fig:recognition}A, which are drawn to six years.} \label{tab:risk} \\
\toprule
Condition & At risk 1 y & 2 y & 5 y & 6 y & 8 y \\
\midrule
\endfirsthead
\caption[]{Participants still at risk just before each horizon in the recognition curves of Figure~\ref{fig:recognition}A, which are drawn to six years.} \\
\toprule
Condition & At risk 1 y & 2 y & 5 y & 6 y & 8 y \\
\midrule
\endhead
\midrule
\multicolumn{6}{r}{Continued on next page} \\
\midrule
\endfoot
\bottomrule
\endlastfoot
Diabetes range & 815 & 577 & 215 & 172 & 37 \\*
CKD range & 501 & 398 & 249 & 213 & 61 \\*
Anemia range & 4,090 & 3,734 & 3,055 & 2,842 & 913 \\
\end{longtable}
}

\subsection{Selective Recording}
{\small\setlength{\tabcolsep}{3.5pt}
\begin{longtable}{lrrrr}
\caption{Held-out prediction of a prior record, among assay-positive participants (the recording propensity) and among participants who are recorded or assay-positive (the PU propensity $c(X)$). The comparator for log loss is a constant estimated in each training fold.} \label{tab:propensity} \\
\toprule
Condition & Participants & N & AUROC & Log-loss improvement \\
\midrule
\endfirsthead
\caption[]{Held-out prediction of a prior record, among assay-positive participants (the recording propensity) and among participants who are recorded or assay-positive (the PU propensity $c(X)$). The comparator for log loss is a constant estimated in each training fold.} \\
\toprule
Condition & Participants & N & AUROC & Log-loss improvement \\
\midrule
\endhead
\midrule
\multicolumn{5}{r}{Continued on next page} \\
\midrule
\endfoot
\bottomrule
\endlastfoot
Diabetes range & Recorded or assay-positive & 10,881 & 0.836 [0.823, 0.849] & 0.087 [0.079, 0.095] \\*
Diabetes range & Assay-positive & 6,663 & 0.875 [0.864, 0.887] & 0.170 [0.158, 0.185] \\*
CKD range & Recorded or assay-positive & 7,032 & 0.703 [0.679, 0.724] & 0.024 [0.019, 0.028] \\*
CKD range & Assay-positive & 1,967 & 0.700 [0.673, 0.726] & 0.062 [0.050, 0.075] \\*
Anemia range & Recorded or assay-positive & 29,011 & 0.630 [0.622, 0.639] & 0.015 [0.013, 0.017] \\*
Anemia range & Assay-positive & 7,600 & 0.670 [0.656, 0.682] & 0.044 [0.038, 0.050] \\
\end{longtable}
}

{\footnotesize\setlength{\tabcolsep}{3.5pt}
\begin{longtable}{lrrrrr}
\caption{Five-year recognition by group among unrecorded assay-positive participants (Aalen--Johansen, death competing), with paired bootstrap differences in percentage points (pp). Comparisons with fewer than 30 participants in a group are omitted. Holm adjustment is within each condition. P-values are bootstrap two-sided tail probabilities with a floor of 0.001.} \label{tab:strata} \\
\toprule
Condition & Comparison & N & 5-y rec.\ (\%) & Diff.\ (pp) [95\% CI] & Holm p \\
\midrule
\endfirsthead
\caption[]{Five-year recognition by group among unrecorded assay-positive participants (Aalen--Johansen, death competing), with paired bootstrap differences in percentage points (pp). Comparisons with fewer than 30 participants in a group are omitted. Holm adjustment is within each condition. P-values are bootstrap two-sided tail probabilities with a floor of 0.001.} \\
\toprule
Condition & Comparison & N & 5-y rec.\ (\%) & Diff.\ (pp) [95\% CI] & Holm p \\
\midrule
\endhead
\midrule
\multicolumn{6}{r}{Continued on next page} \\
\midrule
\endfoot
\bottomrule
\endlastfoot
Diabetes range & Women vs.\ men & 473 vs.\ 771 & 78 vs.\ 83 & $-$5.4 [$-$10.0, $-$0.9] & 0.150 \\
Diabetes range & Most vs.\ least deprived & 370 vs.\ 191 & 83 vs.\ 76 & 6.5 [$-$0.6, 13.8] & 0.224 \\
Diabetes range & Non-white vs.\ white & 157 vs.\ 1,077 & 75 vs.\ 82 & $-$6.6 [$-$13.6, 0.5] & 0.224 \\
Diabetes range & Age 65+ vs.\ $<$55 & 315 vs.\ 325 & 82 vs.\ 83 & $-$1.4 [$-$7.4, 4.5] & 0.604 \\
Diabetes range & Prior-year testing, yes vs.\ no & 59 vs.\ 1,185 & 90 vs.\ 81 & 8.9 [$-$0.4, 16.1] & 0.224 \\
CKD range & Women vs.\ men & 340 vs.\ 288 & 54 vs.\ 56 & $-$2.8 [$-$10.5, 5.3] & 0.956 \\
CKD range & Most vs.\ least deprived & 161 vs.\ 103 & 60 vs.\ 49 & 11.0 [$-$1.6, 23.8] & 0.300 \\
CKD range & Age 65+ vs.\ $<$55 & 357 vs.\ 48 & 56 vs.\ 55 & 0.4 [$-$15.2, 16.3] & 0.998 \\
CKD range & Prior-year testing, yes vs.\ no & 348 vs.\ 280 & 60 vs.\ 48 & 12.0 [4.1, 20.2] & 0.008 \\
Anemia range & Women vs.\ men & 3,135 vs.\ 1,551 & 30 vs.\ 33 & $-$2.8 [$-$5.5, 0.0] & 0.054 \\
Anemia range & Most vs.\ least deprived & 964 vs.\ 857 & 36 vs.\ 25 & 11.2 [6.8, 15.4] & 0.005 \\
Anemia range & Non-white vs.\ white & 349 vs.\ 4,311 & 44 vs.\ 30 & 13.8 [8.4, 19.7] & 0.005 \\
Anemia range & Age 65+ vs.\ $<$55 & 1,013 vs.\ 2,075 & 36 vs.\ 29 & 6.6 [3.0, 10.0] & 0.005 \\*
Anemia range & Prior-year testing, yes vs.\ no & 1,287 vs.\ 3,399 & 38 vs.\ 29 & 9.1 [6.2, 12.3] & 0.005 \\
\end{longtable}
}

{\small\setlength{\tabcolsep}{3.5pt}
\begin{longtable}{lrrrrr}
\caption{Fine--Gray subdistribution hazard ratios for recognition, fitted on the same participants without and with a three-degree-of-freedom spline of severity (HbA1c; creatinine and cystatin C eGFR; hemoglobin deficit below the sex-specific threshold). Both models adjust for age, sex, deprivation quintile, ethnicity, GP data supplier, enrollment year, and prior-year testing, with standard errors clustered by participant. Holm adjustment is over the four comparisons within each model. Hazard ratios are not collapsible, so differences between columns are read qualitatively.} \label{tab:severity} \\
\toprule
Condition & Comparison & HR [95\% CI] & Holm p & With severity, HR [95\% CI] & Holm p \\
\midrule
\endfirsthead
\caption[]{Fine--Gray subdistribution hazard ratios for recognition, fitted on the same participants without and with a three-degree-of-freedom spline of severity (HbA1c; creatinine and cystatin C eGFR; hemoglobin deficit below the sex-specific threshold). Both models adjust for age, sex, deprivation quintile, ethnicity, GP data supplier, enrollment year, and prior-year testing, with standard errors clustered by participant. Holm adjustment is over the four comparisons within each model. Hazard ratios are not collapsible, so differences between columns are read qualitatively.} \\
\toprule
Condition & Comparison & HR [95\% CI] & Holm p & With severity, HR [95\% CI] & Holm p \\
\midrule
\endhead
\midrule
\multicolumn{6}{r}{Continued on next page} \\
\midrule
\endfoot
\bottomrule
\endlastfoot
Diabetes range & Women vs.\ men & 0.88 [0.78, 1.00] & 0.0521 & 0.95 [0.83, 1.08] & 0.413 \\
Diabetes range & Most vs.\ least deprived & 1.30 [1.07, 1.59] & 0.0292 & 1.36 [1.10, 1.66] & 0.0109 \\
Diabetes range & Non-white vs.\ white & 0.80 [0.67, 0.97] & 0.0454 & 0.88 [0.73, 1.05] & 0.314 \\
Diabetes range & Prior-year testing & 1.58 [1.25, 2.01] & $6.16\times10^{-4}$ & 1.84 [1.44, 2.34] & $3.21\times10^{-6}$ \\
CKD range & Women vs.\ men & 0.93 [0.76, 1.14] & 1 & 0.97 [0.79, 1.20] & 1 \\
CKD range & Most vs.\ least deprived & 1.20 [0.87, 1.66] & 0.768 & 1.20 [0.87, 1.65] & 0.823 \\
CKD range & Non-white vs.\ white & 0.99 [0.54, 1.82] & 1 & 0.92 [0.51, 1.69] & 1 \\
CKD range & Prior-year testing & 1.32 [1.08, 1.62] & 0.0265 & 1.23 [1.00, 1.52] & 0.183 \\
Anemia range & Women vs.\ men & 0.90 [0.82, 1.00] & 0.0481 & 0.97 [0.87, 1.08] & 0.554 \\
Anemia range & Most vs.\ least deprived & 1.42 [1.22, 1.65] & $1.88\times10^{-5}$ & 1.41 [1.21, 1.65] & $3.53\times10^{-5}$ \\
Anemia range & Non-white vs.\ white & 1.50 [1.28, 1.76] & $3.15\times10^{-6}$ & 1.56 [1.33, 1.83] & $2.22\times10^{-7}$ \\*
Anemia range & Prior-year testing & 1.24 [1.12, 1.38] & $8.46\times10^{-5}$ & 1.27 [1.14, 1.42] & $3.53\times10^{-5}$ \\
\end{longtable}
}

{\footnotesize\setlength{\tabcolsep}{3.5pt}
\begin{longtable}{lrrrrrr}
\caption{Severity among unrecorded assay-positive participants by group. Groups with fewer than 30 participants are omitted. Mild denotes HbA1c below 53 mmol/mol, creatinine eGFR of 45 or more, or a hemoglobin (Hb) deficit below 1 g/dL. Creatinine eGFR is in mL/min/1.73 m$^2$. Q1 is the least and Q5 the most deprived Townsend quintile.} \label{tab:severitygroups} \\
\toprule
Condition & Severity scale & Grouping & Level & N & Median [IQR] & Mild \\
\midrule
\endfirsthead
\caption[]{Severity among unrecorded assay-positive participants by group. Groups with fewer than 30 participants are omitted. Mild denotes HbA1c below 53 mmol/mol, creatinine eGFR of 45 or more, or a hemoglobin (Hb) deficit below 1 g/dL. Creatinine eGFR is in mL/min/1.73 m$^2$. Q1 is the least and Q5 the most deprived Townsend quintile.} \\
\toprule
Condition & Severity scale & Grouping & Level & N & Median [IQR] & Mild \\
\midrule
\endhead
\midrule
\multicolumn{7}{r}{Continued on next page} \\
\midrule
\endfoot
\bottomrule
\endlastfoot
Diabetes range & HbA1c (mmol/mol) & Deprivation & Q1 & 191 & 53.4 [49.7, 63.3] & 47\% \\
Diabetes range & HbA1c (mmol/mol) & Deprivation & Q5 & 370 & 52.7 [49.5, 61.0] & 52\% \\
Diabetes range & HbA1c (mmol/mol) & Sex & Women & 473 & 51.5 [49.5, 58.1] & 59\% \\
Diabetes range & HbA1c (mmol/mol) & Sex & Men & 771 & 53.6 [49.7, 63.8] & 48\% \\
Diabetes range & HbA1c (mmol/mol) & Ethnicity & White & 1,077 & 52.8 [49.6, 62.4] & 51\% \\
Diabetes range & HbA1c (mmol/mol) & Ethnicity & Non-white & 157 & 51.9 [49.4, 57.6] & 59\% \\
Diabetes range & HbA1c (mmol/mol) & Prior-year testing & No & 1,185 & 52.8 [49.6, 62.2] & 51\% \\
Diabetes range & HbA1c (mmol/mol) & Prior-year testing & Yes & 59 & 50.2 [49.2, 55.0] & 71\% \\
CKD range & Creatinine eGFR & Deprivation & Q1 & 103 & 55.1 [49.6, 57.7] & 85\% \\
CKD range & Creatinine eGFR & Deprivation & Q5 & 161 & 54.6 [49.5, 58.0] & 86\% \\
CKD range & Creatinine eGFR & Sex & Women & 340 & 54.4 [49.7, 58.0] & 87\% \\
CKD range & Creatinine eGFR & Sex & Men & 288 & 54.5 [49.5, 57.6] & 88\% \\
CKD range & Creatinine eGFR & Ethnicity & White & 606 & 54.6 [49.6, 57.8] & 88\% \\
CKD range & Creatinine eGFR & Prior-year testing & No & 280 & 54.6 [50.4, 57.9] & 89\% \\
CKD range & Creatinine eGFR & Prior-year testing & Yes & 348 & 54.2 [49.0, 57.6] & 86\% \\
Anemia range & Hb deficit (g/dL) & Deprivation & Q1 & 857 & 0.4 [0.2, 0.8] & 82\% \\
Anemia range & Hb deficit (g/dL) & Deprivation & Q5 & 964 & 0.4 [0.2, 0.9] & 80\% \\
Anemia range & Hb deficit (g/dL) & Sex & Women & 3,135 & 0.4 [0.2, 0.8] & 81\% \\
Anemia range & Hb deficit (g/dL) & Sex & Men & 1,551 & 0.4 [0.2, 0.9] & 79\% \\
Anemia range & Hb deficit (g/dL) & Ethnicity & White & 4,311 & 0.4 [0.2, 0.8] & 81\% \\
Anemia range & Hb deficit (g/dL) & Ethnicity & Non-white & 349 & 0.4 [0.2, 0.8] & 81\% \\
Anemia range & Hb deficit (g/dL) & Prior-year testing & No & 3,399 & 0.4 [0.2, 0.8] & 80\% \\*
Anemia range & Hb deficit (g/dL) & Prior-year testing & Yes & 1,287 & 0.4 [0.1, 0.7] & 83\% \\
\end{longtable}
}

{\footnotesize\setlength{\tabcolsep}{3.5pt}
\begin{longtable}{lrrrr}
\caption{Anemia without hemoglobinopathy evidence (ICD-10 D56--D58 and primary-care codes that map only to them). Ethnicity and Deprivation are the differences in five-year recognition, in percentage points (pp), between non-white and white participants and between the most and least deprived quintiles, with 95\% bootstrap intervals.} \label{tab:hbp} \\
\toprule
Anemia definition & Unrecorded & 5-y rec.\ (\%) & Ethnicity (pp) & Deprivation (pp) \\
\midrule
\endfirsthead
\caption[]{Anemia without hemoglobinopathy evidence (ICD-10 D56--D58 and primary-care codes that map only to them). Ethnicity and Deprivation are the differences in five-year recognition, in percentage points (pp), between non-white and white participants and between the most and least deprived quintiles, with 95\% bootstrap intervals.} \\
\toprule
Anemia definition & Unrecorded & 5-y rec.\ (\%) & Ethnicity (pp) & Deprivation (pp) \\
\midrule
\endhead
\midrule
\multicolumn{5}{r}{Continued on next page} \\
\midrule
\endfoot
\bottomrule
\endlastfoot
Primary & 4,694 (62\%) & 31 [30, 32] & 13.8 [8.4, 19.7] & 11.2 [6.8, 15.4] \\*
Hemoglobinopathy codes removed & 4,740 (62\%) & 31 [30, 32] & 11.1 [5.8, 16.3] & 10.9 [6.8, 14.9] \\*
Hemoglobinopathy evidence excluded & 4,676 (63\%) & 31 [30, 32] & 12.2 [6.5, 17.5] & 10.9 [6.6, 15.1] \\
\end{longtable}
}

{\small\setlength{\tabcolsep}{3.5pt}
\begin{longtable}{lrrrr}
\caption{Prior recording among assay-positive participants, with Wilson intervals. Q1 is the least and Q5 the most deprived Townsend quintile. These associations do not show whether contact with care precedes or follows a diagnosis.} \label{tab:recordinggroups} \\
\toprule
Condition & Grouping & Level & Assay-positive & With a prior record (\%) [95\% CI] \\
\midrule
\endfirsthead
\caption[]{Prior recording among assay-positive participants, with Wilson intervals. Q1 is the least and Q5 the most deprived Townsend quintile. These associations do not show whether contact with care precedes or follows a diagnosis.} \\
\toprule
Condition & Grouping & Level & Assay-positive & With a prior record (\%) [95\% CI] \\
\midrule
\endhead
\midrule
\multicolumn{5}{r}{Continued on next page} \\
\midrule
\endfoot
\bottomrule
\endlastfoot
Diabetes range & Sex & Women & 2,492 & 81.0 [79.4, 82.5] \\
Diabetes range & Sex & Men & 4,171 & 81.5 [80.3, 82.6] \\
Diabetes range & Deprivation & Q1 & 981 & 80.4 [77.8, 82.8] \\
Diabetes range & Deprivation & Q5 & 1,931 & 80.8 [79.0, 82.5] \\
Diabetes range & Ethnicity & White & 5,893 & 81.7 [80.7, 82.7] \\
Diabetes range & Ethnicity & Non-white & 727 & 78.4 [75.3, 81.2] \\
Diabetes range & Prior-year testing & Yes & 4,715 & 98.7 [98.4, 99.0] \\
Diabetes range & Prior-year testing & No & 1,948 & 39.1 [37.0, 41.3] \\
CKD range & Sex & Women & 943 & 63.9 [60.8, 66.9] \\
CKD range & Sex & Men & 1,024 & 71.9 [69.0, 74.5] \\
CKD range & Deprivation & Q1 & 302 & 65.9 [60.4, 71.0] \\
CKD range & Deprivation & Q5 & 523 & 69.2 [65.1, 73.0] \\
CKD range & Ethnicity & White & 1,855 & 67.3 [65.2, 69.4] \\
CKD range & Ethnicity & Non-white & 102 & 79.4 [70.6, 86.1] \\
CKD range & Prior-year testing & Yes & 1,495 & 76.7 [74.5, 78.8] \\
CKD range & Prior-year testing & No & 472 & 40.7 [36.3, 45.2] \\
Anemia range & Sex & Women & 5,259 & 40.3 [39.0, 41.6] \\
Anemia range & Sex & Men & 2,341 & 33.6 [31.7, 35.5] \\
Anemia range & Deprivation & Q1 & 1,298 & 33.9 [31.4, 36.5] \\
Anemia range & Deprivation & Q5 & 1,760 & 45.1 [42.8, 47.4] \\
Anemia range & Ethnicity & White & 6,741 & 35.9 [34.8, 37.1] \\
Anemia range & Ethnicity & Non-white & 811 & 57.0 [53.5, 60.3] \\
Anemia range & Prior-year testing & Yes & 2,812 & 54.2 [52.4, 56.0] \\*
Anemia range & Prior-year testing & No & 4,788 & 28.9 [27.6, 30.2] \\
\end{longtable}
}

\subsection{Incident Labels and Persistence}
{\small\setlength{\tabcolsep}{3.5pt}
\begin{longtable}{lrrrrr}
\caption{Enrollment value of participants first recognized in each window after enrollment, among study-cohort participants without a prior record. The kidney rows classify creatinine eGFR alone ($<$60, 60--89, or $\geq$90). Intermediate HbA1c is 42--47.9 mmol/mol; anemia has no intermediate state.} \label{tab:composition} \\
\toprule
Recognized condition & Years to recognition & Recognitions & Disease range at enrollment & Intermediate & Normal \\
\midrule
\endfirsthead
\caption[]{Enrollment value of participants first recognized in each window after enrollment, among study-cohort participants without a prior record. The kidney rows classify creatinine eGFR alone ($<$60, 60--89, or $\geq$90). Intermediate HbA1c is 42--47.9 mmol/mol; anemia has no intermediate state.} \\
\toprule
Recognized condition & Years to recognition & Recognitions & Disease range at enrollment & Intermediate & Normal \\
\midrule
\endhead
\midrule
\multicolumn{6}{r}{Continued on next page} \\
\midrule
\endfoot
\bottomrule
\endlastfoot
Diabetes & 0--1 & 866 & 48\% & 26\% & 25\% \\
Diabetes & 1--2 & 798 & 29\% & 32\% & 39\% \\
Diabetes & 2--5 & 2,710 & 13\% & 34\% & 54\% \\
Diabetes & 5--8 & 2,319 & 4\% & 28\% & 68\% \\
CKD & 0--1 & 1,121 & 15\% & 69\% & 17\% \\
CKD & 1--2 & 988 & 12\% & 69\% & 19\% \\
CKD & 2--5 & 2,299 & 8\% & 71\% & 21\% \\
CKD & 5--8 & 1,932 & 4\% & 64\% & 32\% \\
Anemia & 0--1 & 1,925 & 29\% & -- & 71\% \\
Anemia & 1--2 & 1,903 & 17\% & -- & 83\% \\
Anemia & 2--5 & 5,361 & 10\% & -- & 90\% \\*
Anemia & 5--8 & 4,448 & 7\% & -- & 93\% \\
\end{longtable}
}

{\small\setlength{\tabcolsep}{3.5pt}
\begin{longtable}{lrrrr}
\caption{Persistence at the 2012--2013 repeat assessment among assay-positive participants without a prior record, using all UK Biobank participants. Persistence is the share still in the disease range among those still unrecorded; for kidney function it refers to creatinine eGFR below 60.} \label{tab:repeat} \\
\toprule
Condition & Remeasured & Still unrecorded & Still in range & Persistence \\
\midrule
\endfirsthead
\caption[]{Persistence at the 2012--2013 repeat assessment among assay-positive participants without a prior record, using all UK Biobank participants. Persistence is the share still in the disease range among those still unrecorded; for kidney function it refers to creatinine eGFR below 60.} \\
\toprule
Condition & Remeasured & Still unrecorded & Still in range & Persistence \\
\midrule
\endhead
\midrule
\multicolumn{5}{r}{Continued on next page} \\
\midrule
\endfoot
\bottomrule
\endlastfoot
Diabetes range & 44 & 23 & 10 & 43\% \\*
CKD range & 66 & 61 & 43 & 70\% \\*
Anemia range & 516 & 452 & 117 & 26\% \\*
Creatinine-only CKD range & 139 & 129 & 56 & 43\% \\
\end{longtable}
}

\subsection{Case Finding}
{\footnotesize\setlength{\tabcolsep}{2.5pt}
\begin{longtable}{l>{\raggedright\arraybackslash}p{3.1cm}lrr}
\caption{Paired differences in the number found per 1,000 tested, with the top 10\% reselected within each bootstrap sample. Averaged differences are taken over tested fractions of 1--20\%. The comparator uses the same model family.} \label{tab:contrasts} \\
\toprule
Condition & Model & Comparator & Diff.\ at 10\% tested & Averaged, 1--20\% \\
\midrule
\endfirsthead
\caption[]{Paired differences in the number found per 1,000 tested, with the top 10\% reselected within each bootstrap sample. Averaged differences are taken over tested fractions of 1--20\%. The comparator uses the same model family.} \\
\toprule
Condition & Model & Comparator & Diff.\ at 10\% tested & Averaged, 1--20\% \\
\midrule
\endhead
\midrule
\multicolumn{5}{r}{Continued on next page} \\
\midrule
\endfoot
\bottomrule
\endlastfoot
Diabetes range & Recorded (logistic) & Measured & $-$4.0 [$-$6.4, $-$1.9] & $-$5.3 [$-$7.7, $-$3.1] \\
Diabetes range & Recorded (logistic) & Combined & $-$2.1 [$-$3.1, $-$1.1] & $-$1.6 [$-$2.2, $-$1.1] \\
Diabetes range & Combined (logistic) & Measured & $-$1.9 [$-$4.1, 0.1] & $-$3.7 [$-$5.8, $-$1.7] \\
Diabetes range & Incident 5 y (logistic) & Measured & 2.1 [0.3, 3.8] & 2.2 [0.4, 3.9] \\
Diabetes range & Recorded, 115 columns (logistic) & Measured, 115 columns & $-$4.8 [$-$6.7, $-$3.0] & -- \\
Diabetes range & Recorded (boosting) & Measured & $-$5.3 [$-$7.3, $-$3.0] & $-$6.4 [$-$8.4, $-$4.4] \\
Diabetes range & Recorded (boosting) & Combined & $-$3.7 [$-$5.1, $-$2.5] & $-$2.6 [$-$3.5, $-$2.1] \\
Diabetes range & Combined (boosting) & Measured & $-$1.6 [$-$3.3, 0.4] & $-$3.8 [$-$5.4, $-$1.9] \\
Diabetes range & Incident 5 y (boosting) & Measured & 0.3 [$-$1.5, 2.0] & 0.3 [$-$1.3, 1.8] \\
Diabetes range & Recorded, 115 columns (boosting) & Measured, 115 columns & $-$3.9 [$-$5.7, $-$1.8] & -- \\
CKD range & Recorded (logistic) & Measured & 0.1 [$-$1.7, 1.8] & $-$0.1 [$-$2.0, 1.7] \\
CKD range & Recorded (logistic) & Combined & $-$0.5 [$-$1.1, 0.1] & $-$0.7 [$-$1.0, $-$0.4] \\
CKD range & Combined (logistic) & Measured & 0.5 [$-$1.2, 2.3] & 0.5 [$-$1.2, 2.2] \\
CKD range & Incident 5 y (logistic) & Measured & 1.3 [$-$0.4, 2.8] & 0.9 [$-$0.7, 2.5] \\
CKD range & Recorded, 115 columns (logistic) & Measured, 115 columns & $-$2.8 [$-$4.5, $-$1.6] & -- \\
CKD range & Recorded (boosting) & Measured & $-$4.0 [$-$5.6, $-$2.6] & $-$5.5 [$-$7.1, $-$4.0] \\
CKD range & Recorded (boosting) & Combined & $-$1.1 [$-$1.7, $-$0.4] & $-$1.2 [$-$1.6, $-$0.7] \\
CKD range & Combined (boosting) & Measured & $-$2.8 [$-$4.4, $-$1.7] & $-$4.3 [$-$5.8, $-$2.9] \\
CKD range & Incident 5 y (boosting) & Measured & $-$1.6 [$-$3.0, $-$0.4] & $-$2.4 [$-$3.7, $-$1.2] \\
CKD range & Recorded, 115 columns (boosting) & Measured, 115 columns & $-$3.1 [$-$4.6, $-$1.8] & -- \\
Anemia range & Recorded (logistic) & Measured & $-$15.7 [$-$19.6, $-$11.7] & $-$18.1 [$-$21.9, $-$14.0] \\
Anemia range & Recorded (logistic) & Combined & $-$5.7 [$-$7.3, $-$3.8] & $-$7.0 [$-$8.2, $-$5.8] \\
Anemia range & Combined (logistic) & Measured & $-$10.1 [$-$13.5, $-$6.2] & $-$11.0 [$-$14.3, $-$7.4] \\
Anemia range & Incident 5 y (logistic) & Measured & $-$12.9 [$-$16.9, $-$8.9] & $-$12.3 [$-$15.9, $-$8.5] \\
Anemia range & Recorded, 115 columns (logistic) & Measured, 115 columns & $-$16.5 [$-$20.2, $-$13.1] & -- \\
Anemia range & Recorded (boosting) & Measured & $-$26.8 [$-$31.1, $-$22.4] & $-$30.8 [$-$34.9, $-$26.5] \\
Anemia range & Recorded (boosting) & Combined & $-$11.6 [$-$14.0, $-$9.2] & $-$11.0 [$-$12.5, $-$9.4] \\
Anemia range & Combined (boosting) & Measured & $-$15.3 [$-$19.2, $-$11.1] & $-$19.8 [$-$23.2, $-$16.2] \\
Anemia range & Incident 5 y (boosting) & Measured & $-$23.9 [$-$28.3, $-$19.9] & $-$24.0 [$-$28.0, $-$20.4] \\
Anemia range & Recorded, 115 columns (boosting) & Measured, 115 columns & $-$22.3 [$-$26.3, $-$18.3] & -- \\
Calibrated HbA1c & Recorded (logistic) & Measured & $-$7.5 [$-$10.5, $-$4.8] & $-$8.4 [$-$11.2, $-$5.6] \\
Calibrated HbA1c & Recorded (logistic) & Combined & $-$3.9 [$-$5.3, $-$2.8] & $-$3.6 [$-$4.4, $-$2.9] \\
Calibrated HbA1c & Combined (logistic) & Measured & $-$3.6 [$-$6.2, $-$1.2] & $-$4.7 [$-$7.3, $-$2.5] \\
Calibrated HbA1c & Incident 5 y (logistic) & Measured & 0.7 [$-$1.4, 2.9] & 1.5 [$-$0.6, 3.3] \\
Calibrated HbA1c & Recorded (boosting) & Measured & $-$8.0 [$-$10.9, $-$5.7] & $-$10.7 [$-$13.2, $-$8.1] \\
Calibrated HbA1c & Recorded (boosting) & Combined & $-$4.7 [$-$6.4, $-$3.3] & $-$4.1 [$-$5.3, $-$3.3] \\
Calibrated HbA1c & Combined (boosting) & Measured & $-$3.3 [$-$5.6, $-$1.1] & $-$6.6 [$-$8.7, $-$4.2] \\
Calibrated HbA1c & Incident 5 y (boosting) & Measured & $-$0.4 [$-$2.7, 1.3] & $-$1.2 [$-$3.0, 0.5] \\
Diabetes, no medication & Recorded (logistic) & Measured & $-$4.2 [$-$6.5, $-$2.2] & $-$5.5 [$-$7.5, $-$3.5] \\
Diabetes, no medication & Recorded (logistic) & Combined & $-$2.1 [$-$2.8, $-$1.1] & $-$1.8 [$-$2.3, $-$1.2] \\
Diabetes, no medication & Combined (logistic) & Measured & $-$2.2 [$-$4.3, $-$0.2] & $-$3.7 [$-$5.6, $-$1.9] \\
Diabetes, no medication & Incident 5 y (logistic) & Measured & 2.1 [0.1, 4.0] & 1.9 [0.1, 3.7] \\
Diabetes, no medication & Recorded (boosting) & Measured & $-$5.1 [$-$7.3, $-$3.1] & $-$6.2 [$-$8.3, $-$4.1] \\
Diabetes, no medication & Recorded (boosting) & Combined & $-$3.5 [$-$4.8, $-$2.3] & $-$2.5 [$-$3.2, $-$1.7] \\
Diabetes, no medication & Combined (boosting) & Measured & $-$1.6 [$-$3.4, 0.1] & $-$3.7 [$-$5.5, $-$1.9] \\
Diabetes, no medication & Incident 5 y (boosting) & Measured & $-$0.4 [$-$1.9, 1.3] & 0.8 [$-$0.7, 2.4] \\
Anemia, no medication & Recorded (logistic) & Measured & $-$23.2 [$-$27.6, $-$19.3] & $-$24.1 [$-$27.8, $-$20.0] \\
Anemia, no medication & Recorded (logistic) & Combined & $-$12.2 [$-$14.6, $-$10.1] & $-$12.8 [$-$14.3, $-$11.2] \\
Anemia, no medication & Combined (logistic) & Measured & $-$11.0 [$-$14.7, $-$7.6] & $-$11.3 [$-$14.5, $-$8.1] \\
Anemia, no medication & Incident 5 y (logistic) & Measured & $-$18.1 [$-$22.4, $-$14.0] & $-$18.2 [$-$21.8, $-$14.1] \\
Anemia, no medication & Recorded (boosting) & Measured & $-$30.3 [$-$34.3, $-$25.6] & $-$34.5 [$-$38.6, $-$30.1] \\
Anemia, no medication & Recorded (boosting) & Combined & $-$14.0 [$-$16.7, $-$11.5] & $-$16.2 [$-$18.1, $-$14.5] \\
Anemia, no medication & Combined (boosting) & Measured & $-$16.3 [$-$19.7, $-$12.1] & $-$18.2 [$-$21.5, $-$14.9] \\*
Anemia, no medication & Incident 5 y (boosting) & Measured & $-$30.3 [$-$34.4, $-$25.3] & $-$31.0 [$-$34.7, $-$26.9] \\
\end{longtable}
}

{\footnotesize\setlength{\tabcolsep}{3.5pt}
\begin{longtable}{lrrrrrr}
\caption{Operating points for boosting among unrecorded participants. Random and Top 10\%: tests per case found with random testing and with the highest-scoring 10\% tested. Top 1\%: cases per 1,000 tests among the highest-scoring 1\%. Half found: share tested to find half of the assay-positive participants. Exp./obs.: expected and observed cases per 1,000 tested in the top 10\%, with expected cases from isotonic recalibration fitted on the other folds.} \label{tab:operating} \\
\toprule
Condition & Label & Random & Top 10\% & Top 1\% & Half found & Exp./obs. \\
\midrule
\endfirsthead
\caption[]{Operating points for boosting among unrecorded participants. Random and Top 10\%: tests per case found with random testing and with the highest-scoring 10\% tested. Top 1\%: cases per 1,000 tests among the highest-scoring 1\%. Half found: share tested to find half of the assay-positive participants. Exp./obs.: expected and observed cases per 1,000 tested in the top 10\%, with expected cases from isotonic recalibration fitted on the other folds.} \\
\toprule
Condition & Label & Random & Top 10\% & Top 1\% & Half found & Exp./obs. \\
\midrule
\endhead
\midrule
\multicolumn{7}{r}{Continued on next page} \\
\midrule
\endfoot
\bottomrule
\endlastfoot
Diabetes range & Recorded & 133 & 66 & 18 & 32\% & 15.5 / 15.2 \\*
Diabetes range & Measured & 133 & 49 & 34 & 26\% & 20.8 / 20.5 \\*
CKD range & Recorded & 269 & 76 & 20 & 17\% & 13.5 / 13.2 \\*
CKD range & Measured & 269 & 58 & 44 & 12\% & 17.5 / 17.1 \\*
Anemia range & Recorded & 33 & 15 & 95 & 30\% & 65.5 / 64.7 \\*
Anemia range & Measured & 33 & 11 & 159 & 21\% & 91.9 / 91.6 \\
\end{longtable}
}

{\small\setlength{\tabcolsep}{3.5pt}
\begin{longtable}{lrrr}
\caption{Net benefit per 1,000 unrecorded participants when testing everyone whose recalibrated risk exceeds the threshold (boosting).} \label{tab:netbenefit} \\
\toprule
Condition & Risk threshold & Recorded & Measured \\
\midrule
\endfirsthead
\caption[]{Net benefit per 1,000 unrecorded participants when testing everyone whose recalibrated risk exceeds the threshold (boosting).} \\
\toprule
Condition & Risk threshold & Recorded & Measured \\
\midrule
\endhead
\midrule
\multicolumn{4}{r}{Continued on next page} \\
\midrule
\endfoot
\bottomrule
\endlastfoot
Diabetes range & 1\% & 0.7 & 1.2 \\*
CKD range & 1\% & 0.3 & 0.7 \\*
Anemia range & 5\% & 1.5 & 4.9 \\
\end{longtable}
}

{\footnotesize\setlength{\tabcolsep}{3.5pt}
\begin{longtable}{lrrrrr}
\caption{Held-out performance of all models among unrecorded participants. Found per 1,000 tested when the highest-scoring 10\% are tested. AUROC and average precision (AP) are scored against the enrollment value. The last column scores five-year recognition with censoring weights, a different target.} \label{tab:allmodels} \\
\toprule
Condition & Training label & Found per 1,000 [95\% CI] & AUROC & AP & AUROC, 5-y rec. \\
\midrule
\endfirsthead
\caption[]{Held-out performance of all models among unrecorded participants. Found per 1,000 tested when the highest-scoring 10\% are tested. AUROC and average precision (AP) are scored against the enrollment value. The last column scores five-year recognition with censoring weights, a different target.} \\
\toprule
Condition & Training label & Found per 1,000 [95\% CI] & AUROC & AP & AUROC, 5-y rec. \\
\midrule
\endhead
\midrule
\multicolumn{6}{r}{Continued on next page} \\
\midrule
\endfoot
\bottomrule
\endlastfoot
Diabetes range & Recorded (logistic) & 14.3 [12.5, 16.0] & 0.577 & 0.011 & 0.603 \\
Diabetes range & Combined (logistic) & 16.4 [14.6, 18.2] & 0.609 & 0.013 & 0.625 \\
Diabetes range & Measured (logistic) & 18.3 [16.3, 20.4] & 0.638 & 0.016 & 0.616 \\
Diabetes range & Recorded, 115 columns (logistic) & 15.7 [13.8, 17.4] & 0.592 & 0.013 & 0.670 \\
Diabetes range & Measured, 115 columns (logistic) & 20.5 [18.4, 22.6] & 0.671 & 0.018 & 0.676 \\
Diabetes range & Recorded (boosting) & 15.2 [13.3, 16.9] & 0.606 & 0.012 & 0.664 \\
Diabetes range & Combined (boosting) & 18.9 [16.8, 21.1] & 0.653 & 0.014 & 0.689 \\
Diabetes range & Measured (boosting) & 20.5 [18.4, 22.5] & 0.680 & 0.017 & 0.692 \\
Diabetes range & Recorded, 115 columns (boosting) & 15.5 [13.7, 17.4] & 0.598 & 0.012 & 0.673 \\
Diabetes range & Measured, 115 columns (boosting) & 19.4 [17.2, 21.4] & 0.673 & 0.016 & 0.684 \\
Diabetes range & Incident 1 y (logistic) & 17.2 [15.2, 19.2] & 0.622 & 0.014 & 0.636 \\
Diabetes range & Incident 1 y (boosting) & 19.0 [17.0, 21.1] & 0.665 & 0.015 & 0.708 \\
Diabetes range & Incident 2 y (logistic) & 18.5 [16.6, 20.7] & 0.652 & 0.015 & 0.674 \\
Diabetes range & Incident 2 y (boosting) & 18.7 [16.9, 20.7] & 0.670 & 0.016 & 0.714 \\
Diabetes range & Incident 5 y (logistic) & 20.3 [18.3, 22.5] & 0.674 & 0.018 & 0.707 \\
Diabetes range & Incident 5 y (boosting) & 20.8 [18.8, 22.8] & 0.677 & 0.018 & 0.720 \\
CKD range & Recorded (logistic) & 12.7 [11.0, 14.3] & 0.729 & 0.015 & 0.724 \\
CKD range & Combined (logistic) & 13.2 [11.4, 15.0] & 0.749 & 0.017 & 0.731 \\
CKD range & Measured (logistic) & 12.6 [11.0, 14.5] & 0.714 & 0.014 & 0.663 \\
CKD range & Recorded, 115 columns (logistic) & 14.3 [12.5, 16.1] & 0.746 & 0.014 & 0.730 \\
CKD range & Measured, 115 columns (logistic) & 17.2 [15.3, 19.3] & 0.809 & 0.027 & 0.715 \\
CKD range & Recorded (boosting) & 13.2 [11.5, 14.9] & 0.743 & 0.012 & 0.723 \\
CKD range & Combined (boosting) & 14.3 [12.6, 16.1] & 0.765 & 0.016 & 0.732 \\
CKD range & Measured (boosting) & 17.1 [15.2, 19.3] & 0.818 & 0.025 & 0.727 \\
CKD range & Recorded, 115 columns (boosting) & 14.3 [12.6, 16.2] & 0.748 & 0.015 & 0.725 \\
CKD range & Measured, 115 columns (boosting) & 17.4 [15.6, 19.6] & 0.814 & 0.023 & 0.725 \\
CKD range & Incident 1 y (logistic) & 10.8 [9.4, 12.5] & 0.655 & 0.010 & 0.690 \\
CKD range & Incident 1 y (boosting) & 14.5 [12.7, 16.5] & 0.777 & 0.015 & 0.747 \\
CKD range & Incident 2 y (logistic) & 12.3 [10.7, 14.0] & 0.719 & 0.013 & 0.726 \\
CKD range & Incident 2 y (boosting) & 14.7 [12.9, 16.8] & 0.783 & 0.016 & 0.753 \\
CKD range & Incident 5 y (logistic) & 13.9 [12.1, 15.7] & 0.759 & 0.016 & 0.748 \\
CKD range & Incident 5 y (boosting) & 15.5 [13.6, 17.5] & 0.796 & 0.018 & 0.760 \\
Anemia range & Recorded (logistic) & 59.7 [55.9, 63.6] & 0.626 & 0.050 & 0.617 \\
Anemia range & Combined (logistic) & 65.3 [61.4, 69.2] & 0.646 & 0.057 & 0.620 \\
Anemia range & Measured (logistic) & 75.4 [71.2, 79.7] & 0.664 & 0.065 & 0.574 \\
Anemia range & Recorded, 115 columns (logistic) & 56.8 [53.0, 60.3] & 0.628 & 0.050 & 0.631 \\
Anemia range & Measured, 115 columns (logistic) & 73.3 [69.2, 77.4] & 0.667 & 0.062 & 0.590 \\
Anemia range & Recorded (boosting) & 64.7 [60.7, 68.7] & 0.655 & 0.054 & 0.642 \\
Anemia range & Combined (boosting) & 76.3 [72.2, 80.7] & 0.678 & 0.062 & 0.648 \\
Anemia range & Measured (boosting) & 91.6 [87.5, 96.1] & 0.710 & 0.081 & 0.634 \\
Anemia range & Recorded, 115 columns (boosting) & 67.4 [63.6, 71.4] & 0.655 & 0.058 & 0.652 \\
Anemia range & Measured, 115 columns (boosting) & 89.7 [85.5, 94.4] & 0.705 & 0.078 & 0.633 \\
Anemia range & Incident 1 y (logistic) & 54.5 [51.3, 58.2] & 0.580 & 0.046 & 0.625 \\
Anemia range & Incident 1 y (boosting) & 65.5 [61.7, 69.6] & 0.637 & 0.058 & 0.681 \\
Anemia range & Incident 2 y (logistic) & 60.8 [57.5, 64.6] & 0.590 & 0.050 & 0.648 \\
Anemia range & Incident 2 y (boosting) & 70.7 [66.5, 74.6] & 0.644 & 0.061 & 0.691 \\
Anemia range & Incident 5 y (logistic) & 62.5 [58.8, 66.6] & 0.586 & 0.053 & 0.666 \\
Anemia range & Incident 5 y (boosting) & 67.7 [63.6, 71.3] & 0.628 & 0.059 & 0.696 \\
Calibrated HbA1c & Recorded (logistic) & 23.2 [20.9, 25.5] & 0.584 & 0.018 & 0.603 \\
Calibrated HbA1c & Combined (logistic) & 27.1 [24.8, 29.5] & 0.622 & 0.021 & 0.635 \\
Calibrated HbA1c & Measured (logistic) & 30.7 [28.3, 33.4] & 0.659 & 0.025 & 0.649 \\
Calibrated HbA1c & Recorded (boosting) & 24.9 [22.4, 27.2] & 0.620 & 0.020 & 0.664 \\
Calibrated HbA1c & Combined (boosting) & 29.6 [27.2, 32.4] & 0.668 & 0.023 & 0.697 \\
Calibrated HbA1c & Measured (boosting) & 32.9 [30.3, 35.9] & 0.695 & 0.029 & 0.703 \\
Calibrated HbA1c & Incident 5 y (logistic) & 31.4 [29.0, 34.2] & 0.680 & 0.027 & 0.707 \\
Calibrated HbA1c & Incident 5 y (boosting) & 32.5 [29.8, 35.2] & 0.685 & 0.029 & 0.720 \\
Diabetes, no medication & Recorded (logistic) & 14.6 [12.8, 16.5] & 0.579 & 0.012 & 0.602 \\
Diabetes, no medication & Combined (logistic) & 16.6 [14.6, 18.6] & 0.611 & 0.013 & 0.626 \\
Diabetes, no medication & Measured (logistic) & 18.8 [16.7, 20.9] & 0.640 & 0.016 & 0.618 \\
Diabetes, no medication & Recorded (boosting) & 15.8 [13.9, 17.8] & 0.614 & 0.012 & 0.667 \\
Diabetes, no medication & Combined (boosting) & 19.3 [17.2, 21.6] & 0.661 & 0.014 & 0.693 \\
Diabetes, no medication & Measured (boosting) & 20.9 [18.7, 23.1] & 0.684 & 0.017 & 0.696 \\
Diabetes, no medication & Incident 5 y (logistic) & 20.8 [18.7, 23.1] & 0.674 & 0.018 & 0.709 \\
Diabetes, no medication & Incident 5 y (boosting) & 20.5 [18.6, 22.8] & 0.680 & 0.018 & 0.722 \\
Anemia, no medication & Recorded (logistic) & 66.9 [62.9, 70.5] & 0.619 & 0.061 & 0.631 \\
Anemia, no medication & Combined (logistic) & 79.1 [75.0, 83.2] & 0.656 & 0.072 & 0.629 \\
Anemia, no medication & Measured (logistic) & 90.0 [85.8, 94.4] & 0.678 & 0.080 & 0.569 \\
Anemia, no medication & Recorded (boosting) & 75.6 [71.7, 79.8] & 0.643 & 0.065 & 0.650 \\
Anemia, no medication & Combined (boosting) & 89.6 [85.3, 93.9] & 0.692 & 0.080 & 0.649 \\
Anemia, no medication & Measured (boosting) & 105.9 [100.9, 110.5] & 0.720 & 0.097 & 0.616 \\
Anemia, no medication & Incident 5 y (logistic) & 72.0 [67.6, 75.6] & 0.569 & 0.061 & 0.682 \\*
Anemia, no medication & Incident 5 y (boosting) & 75.6 [71.8, 79.8] & 0.592 & 0.066 & 0.697 \\
\end{longtable}
}

{\footnotesize\setlength{\tabcolsep}{3pt}
\begin{longtable}{lrrrrrr}
\caption{Results by GP data supplier. Models are trained on all suppliers, and the top 10\% are selected within each supplier and fold (boosting, found per 1,000 tested).} \label{tab:provider} \\
\toprule
Condition & GP data supplier & Assay-positive & Unrecorded & 5-y rec.\ (\%) & Recorded vs.\ measured & Difference [95\% CI] \\
\midrule
\endfirsthead
\caption[]{Results by GP data supplier. Models are trained on all suppliers, and the top 10\% are selected within each supplier and fold (boosting, found per 1,000 tested).} \\
\toprule
Condition & GP data supplier & Assay-positive & Unrecorded & 5-y rec.\ (\%) & Recorded vs.\ measured & Difference [95\% CI] \\
\midrule
\endhead
\midrule
\multicolumn{7}{r}{Continued on next page} \\
\midrule
\endfoot
\bottomrule
\endlastfoot
Diabetes range & England (TPP) & 4,843 & 18\% & 82 [80, 85] & 15.4 vs.\ 21.8 & $-$6.4 [$-$8.8, $-$3.7] \\
Diabetes range & England (Vision) & 418 & 21\% & 82 [74, 90] & 12.6 vs.\ 16.6 & $-$4.0 [$-$11.9, 4.7] \\
Diabetes range & Scotland & 758 & 21\% & 75 [69, 82] & 20.6 vs.\ 19.6 & 1.0 [$-$5.5, 7.0] \\
Diabetes range & Wales & 644 & 17\% & 81 [73, 88] & 11.0 vs.\ 15.6 & $-$4.5 [$-$9.7, 0.6] \\
CKD range & England (TPP) & 1,403 & 30\% & 56 [51, 61] & 13.1 vs.\ 16.9 & $-$3.9 [$-$5.7, $-$2.3] \\
CKD range & England (Vision) & 128 & 29\% & 57 [41, 73] & 7.8 vs.\ 11.7 & $-$3.9 [$-$8.6, 1.6] \\
CKD range & Scotland & 242 & 41\% & 54 [44, 64] & 15.0 vs.\ 18.4 & $-$3.4 [$-$7.8, 1.5] \\
CKD range & Wales & 194 & 38\% & 50 [38, 62] & 18.8 vs.\ 20.1 & $-$1.3 [$-$6.3, 2.5] \\
Anemia range & England (TPP) & 5,554 & 60\% & 33 [31, 35] & 67.4 vs.\ 93.6 & $-$26.2 [$-$31.3, $-$21.1] \\
Anemia range & England (Vision) & 498 & 60\% & 27 [22, 33] & 58.7 vs.\ 64.5 & $-$5.8 [$-$20.7, 9.1] \\
Anemia range & Scotland & 865 & 72\% & 21 [17, 24] & 66.2 vs.\ 91.7 & $-$25.5 [$-$40.1, $-$13.0] \\*
Anemia range & Wales & 683 & 65\% & 33 [29, 37] & 61.7 vs.\ 91.2 & $-$29.5 [$-$43.2, $-$14.4] \\
\end{longtable}
}

{\footnotesize\setlength{\tabcolsep}{3.5pt}
\begin{longtable}{lrrrrrr}
\caption{Stricter thresholds. The models are those of the primary analysis, with the measured label at the primary threshold, and are scored against the stricter value in the same held-out unrecorded participants (boosting, top 10\% tested). Kidney function at the stricter threshold has too few unrecorded participants for case finding. Recorded vs.\ measured is the number found per 1,000 tested with each label, and Fewer is one minus the ratio of recorded to measured.} \label{tab:strict} \\
\toprule
Condition & Threshold & Assay-positive & Unrecorded & 5-y rec.\ (\%) & Recorded vs.\ measured & Fewer \\
\midrule
\endfirsthead
\caption[]{Stricter thresholds. The models are those of the primary analysis, with the measured label at the primary threshold, and are scored against the stricter value in the same held-out unrecorded participants (boosting, top 10\% tested). Kidney function at the stricter threshold has too few unrecorded participants for case finding. Recorded vs.\ measured is the number found per 1,000 tested with each label, and Fewer is one minus the ratio of recorded to measured.} \\
\toprule
Condition & Threshold & Assay-positive & Unrecorded & 5-y rec.\ (\%) & Recorded vs.\ measured & Fewer \\
\midrule
\endhead
\midrule
\multicolumn{7}{r}{Continued on next page} \\
\midrule
\endfoot
\bottomrule
\endlastfoot
Diabetes range & Primary threshold & 6,663 & 1,245 (19\%) & 81 [79, 83] & 15.2 vs.\ 20.5 & 26\% \\*
Diabetes range & HbA1c $\geq$53 & 4,459 & 598 (13\%) & 91 [89, 93] & 5.6 vs.\ 8.2 & 32\% \\*
Anemia range & Primary threshold & 7,600 & 4,694 (62\%) & 31 [30, 32] & 64.7 vs.\ 91.6 & 29\% \\*
Anemia range & Hemoglobin $<$11 & 1,311 & 657 (50\%) & 50 [46, 53] & 9.7 vs.\ 18.5 & 48\% \\*
CKD range & Primary threshold & 1,967 & 628 (32\%) & 55 [51, 59] & 13.2 vs.\ 17.1 & 23\% \\*
CKD range & Both eGFR $<$45 & 494 & 65 (13\%) & 57 [44, 69] & -- & -- \\
\end{longtable}
}

{\footnotesize\setlength{\tabcolsep}{3.5pt}
\begin{longtable}{lrrrrr}
\caption{Feature sets. The 627 columns are the full representation, the 115 columns drop the code history, and the 67 columns keep only demographic, GP data supplier, assessment-center, calendar, and registration information. Recorded vs.\ measured is the number found per 1,000 tested (top 10\%) with each label, and Fewer is one minus the ratio of recorded to measured.} \label{tab:representation} \\
\toprule
Condition & Columns & Model & Recorded vs.\ measured & Difference [95\% CI] & Fewer \\
\midrule
\endfirsthead
\caption[]{Feature sets. The 627 columns are the full representation, the 115 columns drop the code history, and the 67 columns keep only demographic, GP data supplier, assessment-center, calendar, and registration information. Recorded vs.\ measured is the number found per 1,000 tested (top 10\%) with each label, and Fewer is one minus the ratio of recorded to measured.} \\
\toprule
Condition & Columns & Model & Recorded vs.\ measured & Difference [95\% CI] & Fewer \\
\midrule
\endhead
\midrule
\multicolumn{6}{r}{Continued on next page} \\
\midrule
\endfoot
\bottomrule
\endlastfoot
Diabetes range & 627 & boosting & 15.2 vs.\ 20.5 & $-$5.3 [$-$7.3, $-$3.0] & 26\% \\
Diabetes range & 115 & boosting & 15.5 vs.\ 19.4 & $-$3.9 [$-$5.7, $-$1.8] & 20\% \\
Diabetes range & 67 & boosting & 16.2 vs.\ 17.6 & $-$1.4 [$-$3.1, 0.5] & 8\% \\
Diabetes range & 627 & logistic & 14.3 vs.\ 18.3 & $-$4.0 [$-$6.4, $-$1.9] & 22\% \\
Diabetes range & 115 & logistic & 15.7 vs.\ 20.5 & $-$4.8 [$-$6.7, $-$3.0] & 23\% \\
Diabetes range & 67 & logistic & 17.6 vs.\ 18.8 & $-$1.1 [$-$2.5, 0.3] & 6\% \\
CKD range & 627 & boosting & 13.2 vs.\ 17.1 & $-$4.0 [$-$5.6, $-$2.6] & 23\% \\
CKD range & 115 & boosting & 14.3 vs.\ 17.4 & $-$3.1 [$-$4.6, $-$1.8] & 18\% \\
CKD range & 67 & boosting & 11.2 vs.\ 12.9 & $-$1.7 [$-$3.4, $-$0.1] & 13\% \\
CKD range & 627 & logistic & 12.7 vs.\ 12.6 & 0.1 [$-$1.7, 1.8] & 0\% \\
CKD range & 115 & logistic & 14.3 vs.\ 17.2 & $-$2.8 [$-$4.5, $-$1.6] & 17\% \\
CKD range & 67 & logistic & 10.8 vs.\ 13.3 & $-$2.4 [$-$4.0, $-$0.7] & 18\% \\
Anemia range & 627 & boosting & 64.7 vs.\ 91.6 & $-$26.8 [$-$31.1, $-$22.4] & 29\% \\
Anemia range & 115 & boosting & 67.4 vs.\ 89.7 & $-$22.3 [$-$26.3, $-$18.3] & 25\% \\
Anemia range & 67 & boosting & 65.2 vs.\ 72.9 & $-$7.7 [$-$10.9, $-$4.1] & 11\% \\
Anemia range & 627 & logistic & 59.7 vs.\ 75.4 & $-$15.7 [$-$19.6, $-$11.7] & 21\% \\
Anemia range & 115 & logistic & 56.8 vs.\ 73.3 & $-$16.5 [$-$20.2, $-$13.1] & 23\% \\*
Anemia range & 67 & logistic & 48.8 vs.\ 65.7 & $-$17.0 [$-$20.9, $-$12.9] & 26\% \\
\end{longtable}
}

{\scriptsize\setlength{\tabcolsep}{3.5pt}
\begin{longtable}{lrrrrrr}
\caption{Groups when the same overall 10\% are tested, recorded-label / measured-label boosting. Selected is the share of the group that is tested; Precision is the share of those tested who are assay-positive; Recall is the share of the group's assay-positive participants who are tested; Recall diff.\ is measured-label minus recorded-label recall in percentage points (pp). Groups with fewer than 30 assay-positive participants are omitted. Intervals are not adjusted for multiplicity.} \label{tab:groups} \\
\toprule
Condition & Group & Assay-positive & Selected & Precision (\%) & Recall & Recall diff.\ (pp) [95\% CI] \\
\midrule
\endfirsthead
\caption[]{Groups when the same overall 10\% are tested, recorded-label / measured-label boosting. Selected is the share of the group that is tested; Precision is the share of those tested who are assay-positive; Recall is the share of the group's assay-positive participants who are tested; Recall diff.\ is measured-label minus recorded-label recall in percentage points (pp). Groups with fewer than 30 assay-positive participants are omitted. Intervals are not adjusted for multiplicity.} \\
\toprule
Condition & Group & Assay-positive & Selected & Precision (\%) & Recall & Recall diff.\ (pp) [95\% CI] \\
\midrule
\endhead
\midrule
\multicolumn{7}{r}{Continued on next page} \\
\midrule
\endfoot
\bottomrule
\endlastfoot
Diabetes range & Women & 473 & 9\% / 5\% & 1.1 / 1.8 & 19\% / 19\% & 0.6 [$-$4.1, 4.8] \\
Diabetes range & Men & 772 & 12\% / 16\% & 1.9 / 2.1 & 21\% / 32\% & 10.9 [7.5, 14.5] \\
Diabetes range & White & 1,078 & 9\% / 7\% & 1.3 / 1.6 & 18\% / 18\% & $-$0.6 [$-$3.3, 2.0] \\
Diabetes range & Non-white & 157 & 26\% / 74\% & 3.2 / 3.1 & 32\% / 89\% & 56.1 [47.1, 63.8] \\
Diabetes range & Least deprived & 192 & 8\% / 7\% & 0.9 / 1.6 & 12\% / 20\% & 7.8 [1.0, 13.8] \\
Diabetes range & Most deprived & 370 & 14\% / 16\% & 2.3 / 3.2 & 25\% / 38\% & 13.0 [7.1, 18.3] \\
Diabetes range & Age $<$55 & 325 & 7\% / 6\% & 1.3 / 2.0 & 16\% / 23\% & 6.5 [1.2, 11.9] \\
Diabetes range & Age 65+ & 315 & 16\% / 17\% & 1.4 / 1.9 & 23\% / 32\% & 8.6 [3.4, 14.5] \\
CKD range & Women & 340 & 9\% / 9\% & 1.4 / 1.7 & 32\% / 43\% & 10.6 [5.4, 15.8] \\
CKD range & Men & 288 & 12\% / 11\% & 1.3 / 1.7 & 39\% / 50\% & 10.8 [5.4, 17.0] \\
CKD range & White & 606 & 10\% / 10\% & 1.3 / 1.7 & 35\% / 46\% & 11.2 [7.5, 15.2] \\
CKD range & Least deprived & 103 & 9\% / 9\% & 1.4 / 1.7 & 41\% / 53\% & 12.6 [3.6, 22.4] \\
CKD range & Most deprived & 161 & 12\% / 12\% & 1.8 / 2.3 & 41\% / 50\% & 8.7 [0.6, 16.5] \\
CKD range & Age $<$55 & 48 & 2\% / 0\% & 0.5 / 1.2 & 12\% / 4\% & $-$8.3 [$-$16.7, $-$1.9] \\
CKD range & Age 65+ & 357 & 25\% / 41\% & 1.9 / 1.8 & 41\% / 64\% & 23.2 [18.6, 28.6] \\
Anemia range & Women & 3,139 & 17\% / 16\% & 6.0 / 8.5 & 26\% / 33\% & 7.9 [5.9, 9.6] \\
Anemia range & Men & 1,555 & 3\% / 4\% & 9.4 / 11.7 & 12\% / 23\% & 10.7 [8.8, 12.6] \\
Anemia range & White & 4,319 & 9\% / 9\% & 5.9 / 8.8 & 18\% / 28\% & 9.3 [7.9, 10.8] \\
Anemia range & Non-white & 349 & 35\% / 29\% & 9.9 / 12.5 & 55\% / 58\% & 3.2 [$-$2.8, 8.6] \\
Anemia range & Least deprived & 858 & 8\% / 8\% & 6.3 / 8.7 & 19\% / 28\% & 8.3 [4.7, 11.5] \\
Anemia range & Most deprived & 966 & 14\% / 13\% & 7.0 / 9.6 & 27\% / 35\% & 8.4 [5.0, 11.9] \\
Anemia range & Age $<$55 & 2,077 & 14\% / 19\% & 6.5 / 8.2 & 25\% / 43\% & 18.4 [15.7, 21.0] \\
Anemia range & Age 65+ & 1,016 & 8\% / 7\% & 7.9 / 12.1 & 19\% / 25\% & 5.6 [3.0, 8.4] \\
Calibrated HbA1c & Women & 788 & 9\% / 7\% & 2.1 / 3.2 & 21\% / 25\% & 4.3 [1.0, 7.7] \\
Calibrated HbA1c & Men & 1,102 & 12\% / 14\% & 2.9 / 3.4 & 22\% / 31\% & 8.9 [6.4, 12.0] \\
Calibrated HbA1c & White & 1,643 & 9\% / 8\% & 2.2 / 2.8 & 20\% / 20\% & 0.9 [$-$1.0, 3.3] \\
Calibrated HbA1c & Non-white & 236 & 26\% / 66\% & 5.5 / 5.0 & 36\% / 84\% & 47.5 [40.5, 54.5] \\
Calibrated HbA1c & Least deprived & 297 & 8\% / 7\% & 1.8 / 2.5 & 17\% / 21\% & 4.4 [$-$0.7, 10.0] \\
Calibrated HbA1c & Most deprived & 544 & 14\% / 16\% & 3.8 / 4.5 & 29\% / 37\% & 8.3 [4.3, 12.9] \\
Calibrated HbA1c & Age $<$55 & 459 & 7\% / 5\% & 1.8 / 3.3 & 16\% / 21\% & 4.8 [0.9, 9.0] \\
Calibrated HbA1c & Age 65+ & 525 & 16\% / 19\% & 2.7 / 3.1 & 26\% / 36\% & 10.1 [6.4, 14.7] \\
Diabetes, no medication & Women & 477 & 8\% / 5\% & 1.1 / 2.0 & 18\% / 21\% & 2.3 [$-$1.8, 6.6] \\
Diabetes, no medication & Men & 774 & 12\% / 16\% & 2.0 / 2.1 & 22\% / 32\% & 9.4 [6.2, 13.0] \\
Diabetes, no medication & White & 1,082 & 9\% / 7\% & 1.3 / 1.7 & 19\% / 18\% & $-$0.1 [$-$2.6, 2.7] \\
Diabetes, no medication & Non-white & 159 & 25\% / 72\% & 3.8 / 3.1 & 36\% / 86\% & 49.7 [41.3, 58.7] \\
Diabetes, no medication & Least deprived & 192 & 8\% / 7\% & 0.9 / 1.4 & 14\% / 19\% & 5.2 [$-$0.6, 12.2] \\
Diabetes, no medication & Most deprived & 374 & 14\% / 16\% & 2.5 / 3.2 & 27\% / 38\% & 11.0 [5.8, 15.8] \\
Diabetes, no medication & Age $<$55 & 327 & 6\% / 6\% & 1.3 / 2.2 & 15\% / 24\% & 9.5 [4.4, 14.7] \\
Diabetes, no medication & Age 65+ & 315 & 16\% / 17\% & 1.5 / 1.9 & 25\% / 32\% & 7.6 [2.1, 13.3] \\
Anemia, no medication & Women & 3,828 & 14\% / 15\% & 7.2 / 10.1 & 23\% / 34\% & 11.1 [9.4, 12.6] \\
Anemia, no medication & Men & 1,704 & 5\% / 4\% & 8.6 / 12.7 & 19\% / 23\% & 3.8 [1.9, 5.5] \\
Anemia, no medication & White & 5,010 & 9\% / 9\% & 6.7 / 9.9 & 18\% / 27\% & 8.7 [7.3, 9.9] \\
Anemia, no medication & Non-white & 493 & 39\% / 39\% & 12.4 / 14.5 & 60\% / 71\% & 11.0 [6.5, 15.2] \\
Anemia, no medication & Least deprived & 1,000 & 8\% / 8\% & 6.5 / 9.9 & 18\% / 27\% & 9.7 [6.5, 12.7] \\
Anemia, no medication & Most deprived & 1,194 & 14\% / 14\% & 9.6 / 11.7 & 32\% / 38\% & 6.5 [3.7, 9.3] \\
Anemia, no medication & Age $<$55 & 2,546 & 9\% / 18\% & 8.7 / 9.4 & 18\% / 41\% & 23.1 [20.9, 25.1] \\*
Anemia, no medication & Age 65+ & 1,150 & 13\% / 8\% & 8.1 / 13.5 & 29\% / 27\% & $-$1.2 [$-$3.6, 1.3] \\
\end{longtable}
}

{\small\setlength{\tabcolsep}{3.5pt}
\begin{longtable}{lrrrr}
\caption{AUROC of the measured-label and five-year incident-label boosting models against two targets in the same held-out unrecorded participants, with paired bootstrap intervals. Recognition is scored with inverse-probability-of-censoring weights, and deaths before recognition count as negatives.} \label{tab:targets} \\
\toprule
Condition & Scored against & Measured & Incident 5 y & Difference \\
\midrule
\endfirsthead
\caption[]{AUROC of the measured-label and five-year incident-label boosting models against two targets in the same held-out unrecorded participants, with paired bootstrap intervals. Recognition is scored with inverse-probability-of-censoring weights, and deaths before recognition count as negatives.} \\
\toprule
Condition & Scored against & Measured & Incident 5 y & Difference \\
\midrule
\endhead
\midrule
\multicolumn{5}{r}{Continued on next page} \\
\midrule
\endfoot
\bottomrule
\endlastfoot
Diabetes range & Enrollment value & 0.680 [0.667, 0.696] & 0.677 [0.663, 0.692] & 0.004 [$-$0.005, 0.014] \\*
Diabetes range & Recognition within 5 y & 0.692 [0.684, 0.700] & 0.720 [0.712, 0.728] & $-$0.028 [$-$0.033, $-$0.023] \\*
CKD range & Enrollment value & 0.818 [0.803, 0.833] & 0.796 [0.779, 0.812] & 0.022 [0.013, 0.031] \\*
CKD range & Recognition within 5 y & 0.727 [0.720, 0.734] & 0.760 [0.753, 0.767] & $-$0.033 [$-$0.037, $-$0.029] \\*
Anemia range & Enrollment value & 0.710 [0.702, 0.718] & 0.628 [0.619, 0.636] & 0.082 [0.074, 0.090] \\*
Anemia range & Recognition within 5 y & 0.634 [0.628, 0.640] & 0.696 [0.690, 0.701] & $-$0.062 [$-$0.067, $-$0.057] \\
\end{longtable}
}

\subsection{Propensity Correction}
{\small\setlength{\tabcolsep}{3.5pt}
\begin{longtable}{lrrr}
\caption{Paired comparisons of PU scores when the highest-scoring 10\% are tested (found per 1,000 tested).} \label{tab:pucontrasts} \\
\toprule
Condition & Score & Comparator & Difference [95\% CI] \\
\midrule
\endfirsthead
\caption[]{Paired comparisons of PU scores when the highest-scoring 10\% are tested (found per 1,000 tested).} \\
\toprule
Condition & Score & Comparator & Difference [95\% CI] \\
\midrule
\endhead
\midrule
\multicolumn{4}{r}{Continued on next page} \\
\midrule
\endfoot
\bottomrule
\endlastfoot
Diabetes range & PU unrecorded (logistic) & Calibrated recorded (logistic) & 2.5 [1.1, 4.2] \\
Diabetes range & PU population (logistic) & Calibrated recorded (logistic) & 0.4 [$-$0.1, 0.9] \\
Diabetes range & PU unrecorded (logistic) & PU population (logistic) & 2.2 [0.7, 3.7] \\
Diabetes range & PU unrecorded (logistic) & Measured, 115 columns (logistic) & $-$2.3 [$-$3.8, $-$0.8] \\
Diabetes range & PU, 10\% assays (logistic) & Measured, 10\% assays (logistic) & $-$0.4 [$-$2.1, 1.6] \\
Diabetes range & PU unrecorded (boosting) & Calibrated recorded (boosting) & 0.4 [$-$1.3, 2.0] \\
Diabetes range & PU population (boosting) & Calibrated recorded (boosting) & 0.3 [$-$0.2, 0.7] \\
Diabetes range & PU unrecorded (boosting) & PU population (boosting) & 0.1 [$-$1.5, 1.7] \\
Diabetes range & PU unrecorded (boosting) & Measured, 115 columns (boosting) & $-$3.3 [$-$5.1, $-$1.4] \\
CKD range & PU unrecorded (logistic) & Calibrated recorded (logistic) & 2.7 [1.5, 3.9] \\
CKD range & PU population (logistic) & Calibrated recorded (logistic) & 0.5 [0.2, 1.1] \\
CKD range & PU unrecorded (logistic) & PU population (logistic) & 2.1 [1.0, 3.2] \\
CKD range & PU unrecorded (logistic) & Measured, 115 columns (logistic) & $-$0.2 [$-$1.2, 0.5] \\
CKD range & PU, 10\% assays (logistic) & Measured, 10\% assays (logistic) & 1.1 [$-$0.4, 2.3] \\
CKD range & PU unrecorded (boosting) & Calibrated recorded (boosting) & 2.3 [1.2, 3.5] \\
CKD range & PU population (boosting) & Calibrated recorded (boosting) & 0.4 [0.0, 0.8] \\
CKD range & PU unrecorded (boosting) & PU population (boosting) & 2.0 [0.8, 3.1] \\
CKD range & PU unrecorded (boosting) & Measured, 115 columns (boosting) & $-$0.8 [$-$2.0, 0.4] \\
Anemia range & PU unrecorded (logistic) & Calibrated recorded (logistic) & 17.3 [14.0, 20.7] \\
Anemia range & PU population (logistic) & Calibrated recorded (logistic) & 2.3 [1.4, 4.0] \\
Anemia range & PU unrecorded (logistic) & PU population (logistic) & 15.0 [11.6, 17.7] \\
Anemia range & PU unrecorded (logistic) & Measured, 115 columns (logistic) & 0.8 [$-$1.0, 2.2] \\
Anemia range & PU, 10\% assays (logistic) & Measured, 10\% assays (logistic) & $-$4.5 [$-$7.5, $-$1.4] \\
Anemia range & PU unrecorded (boosting) & Calibrated recorded (boosting) & 13.6 [10.3, 17.2] \\
Anemia range & PU population (boosting) & Calibrated recorded (boosting) & 3.1 [1.6, 4.8] \\
Anemia range & PU unrecorded (boosting) & PU population (boosting) & 10.5 [7.5, 13.7] \\*
Anemia range & PU unrecorded (boosting) & Measured, 115 columns (boosting) & $-$9.0 [$-$12.5, $-$5.3] \\
\end{longtable}
}

{\small\setlength{\tabcolsep}{3.5pt}
\begin{longtable}{lrrr}
\caption{PU correction on the 115-column representation. Calibrated recorded ranks by $g$, PU population by $p=g/c$, and PU unrecorded by $(p-g)/(1-g)$, the probability of a case among participants with $R=0$; all three use the same $g$ model (Appendix~\ref{app:pu}). The 10\%-assay comparisons use the same randomly chosen assay subset.} \label{tab:pu} \\
\toprule
Condition & Score & Found per 1,000 [95\% CI] & AUROC \\
\midrule
\endfirsthead
\caption[]{PU correction on the 115-column representation. Calibrated recorded ranks by $g$, PU population by $p=g/c$, and PU unrecorded by $(p-g)/(1-g)$, the probability of a case among participants with $R=0$; all three use the same $g$ model (Appendix~\ref{app:pu}). The 10\%-assay comparisons use the same randomly chosen assay subset.} \\
\toprule
Condition & Score & Found per 1,000 [95\% CI] & AUROC \\
\midrule
\endhead
\midrule
\multicolumn{4}{r}{Continued on next page} \\
\midrule
\endfoot
\bottomrule
\endlastfoot
Diabetes range & Calibrated recorded (logistic) & 15.6 [13.7, 17.5] & 0.592 \\
Diabetes range & PU population (logistic) & 16.0 [14.2, 17.8] & 0.604 \\
Diabetes range & PU unrecorded (logistic) & 18.2 [16.3, 20.2] & 0.632 \\
Diabetes range & PU floor 0.005 (logistic) & 18.2 [16.3, 20.2] & 0.632 \\
Diabetes range & Calibrated recorded (boosting) & 15.7 [14.0, 17.6] & 0.599 \\
Diabetes range & PU population (boosting) & 16.0 [14.1, 17.7] & 0.626 \\
Diabetes range & PU unrecorded (boosting) & 16.1 [14.3, 18.1] & 0.623 \\
Diabetes range & PU floor 0.005 (boosting) & 16.1 [14.3, 18.1] & 0.623 \\
Diabetes range & PU, 10\% assays (logistic) & 16.4 [14.6, 18.4] & 0.597 \\
Diabetes range & Measured, 10\% assays (logistic) & 16.8 [14.8, 18.9] & 0.605 \\
CKD range & Calibrated recorded (logistic) & 14.3 [12.4, 16.2] & 0.746 \\
CKD range & PU population (logistic) & 14.8 [13.0, 16.8] & 0.757 \\
CKD range & PU unrecorded (logistic) & 16.9 [14.9, 18.9] & 0.802 \\
CKD range & PU floor 0.005 (logistic) & 16.9 [14.9, 18.9] & 0.802 \\
CKD range & Calibrated recorded (boosting) & 14.3 [12.5, 16.1] & 0.748 \\
CKD range & PU population (boosting) & 14.7 [12.9, 16.6] & 0.759 \\
CKD range & PU unrecorded (boosting) & 16.6 [14.7, 18.7] & 0.798 \\
CKD range & PU floor 0.005 (boosting) & 16.6 [14.7, 18.7] & 0.798 \\
CKD range & PU, 10\% assays (logistic) & 15.2 [13.2, 17.2] & 0.764 \\
CKD range & Measured, 10\% assays (logistic) & 14.2 [12.5, 15.9] & 0.763 \\
Anemia range & Calibrated recorded (logistic) & 56.8 [53.0, 60.1] & 0.628 \\
Anemia range & PU population (logistic) & 59.2 [55.5, 62.7] & 0.639 \\
Anemia range & PU unrecorded (logistic) & 74.1 [69.8, 78.2] & 0.667 \\
Anemia range & PU floor 0.005 (logistic) & 74.1 [69.8, 78.2] & 0.667 \\
Anemia range & Calibrated recorded (boosting) & 67.1 [63.1, 71.0] & 0.655 \\
Anemia range & PU population (boosting) & 70.3 [66.1, 74.6] & 0.666 \\
Anemia range & PU unrecorded (boosting) & 80.7 [76.2, 85.2] & 0.690 \\
Anemia range & PU floor 0.005 (boosting) & 80.7 [76.2, 85.2] & 0.690 \\
Anemia range & PU, 10\% assays (logistic) & 61.6 [57.7, 65.4] & 0.644 \\*
Anemia range & Measured, 10\% assays (logistic) & 66.1 [62.4, 70.1] & 0.644 \\
\end{longtable}
}

\subsection{Composition of Anemia-Range Participants}
{\small\setlength{\tabcolsep}{3.5pt}
\begin{longtable}{lrrr}
\caption{Anemia-range participants by pregnancy status, mean corpuscular volume (MCV), menopause, and hemoglobinopathy evidence.} \label{tab:anemiagroups} \\
\toprule
Anemia-range participants & Assay-positive & Unrecorded & 5-y rec.\ (\%) [95\% CI] \\
\midrule
\endfirsthead
\caption[]{Anemia-range participants by pregnancy status, mean corpuscular volume (MCV), menopause, and hemoglobinopathy evidence.} \\
\toprule
Anemia-range participants & Assay-positive & Unrecorded & 5-y rec.\ (\%) [95\% CI] \\
\midrule
\endhead
\midrule
\multicolumn{4}{r}{Continued on next page} \\
\midrule
\endfoot
\bottomrule
\endlastfoot
All & 7,600 & 62\% & 31 [30, 32] \\
Pregnant or unsure at enrollment & 19 & 79\% & n $<$ 30 \\
Excluding pregnant or unsure & 7,581 & 62\% & 31 [30, 32] \\
MCV $<$80 fL & 1,413 & 48\% & 56 [52, 59] \\
MCV 80--100 fL & 5,939 & 65\% & 27 [25, 28] \\
MCV $>$100 fL & 246 & 59\% & 31 [24, 39] \\
Women, premenopausal & 2,294 & 59\% & 35 [32, 37] \\
Women, postmenopausal & 2,268 & 62\% & 26 [23, 28] \\
Women, hysterectomy or unsure & 670 & 57\% & 30 [25, 34] \\
Women, age $<$50 & 2,077 & 58\% & 35 [32, 37] \\
Women, age $\geq$50 & 3,182 & 61\% & 27 [25, 29] \\
Men & 2,341 & 66\% & 33 [31, 35] \\
Any hemoglobinopathy evidence & 175 & 10\% & n $<$ 30 \\*
Hemoglobinopathy evidence before enrollment & 129 & 0\% & n $<$ 30 \\
\end{longtable}
}

\end{document}